\documentclass[journal,twoside,web]{ieeecolor}
\usepackage{tmi}
\usepackage{cite}
\usepackage{amsmath,amssymb,amsfonts}
\usepackage{graphicx}
\usepackage{textcomp}
\usepackage{algpseudocode}
\usepackage{multirow}
\usepackage{epstopdf}
\usepackage{amsmath}
\usepackage{lipsum} 
\usepackage{float} 
\usepackage{subcaption}
\usepackage[utf8]{inputenc}
\usepackage{mdframed}
\usepackage[ruled,boxed]{algorithm2e}
\usepackage{xcolor}
\usepackage[table]{xcolor}
\usepackage{color}
\definecolor{bottomcolor}{RGB}{247, 248, 236}
\usepackage{threeparttable}
\usepackage[pagebackref=false,breaklinks=true,colorlinks=true,citecolor=citeblue,bookmarks=false]{hyperref}
\usepackage{caption}
\usepackage{placeins}  
\usepackage{booktabs} 
\usepackage{multirow}
\usepackage{url}

\definecolor{grey}{RGB}{235,235,230}
\definecolor{orange}{RGB}{255,244,213}
\definecolor{purple}{RGB}{235,222,240}
\definecolor{green}{RGB}{199,230,204}
\definecolor{brown}{RGB}{222,207,186}
 
\newcommand{\linepurple}[1]{\rowcolor{purple}\cellcolor{white}#1}
\newcommand{\lineorange}[1]{\cellcolor{orange}#1}
\newcommand{\linegreen}[1]{\cellcolor{green}{#1}}
\newcommand{\linegrey}[1]{\cellcolor{grey}#1}

\definecolor{revmaroon}{RGB}{128,0,0} 
\newcommand{\boldres}[1]{{\textbf{\textcolor{revmaroon}{#1}}}}
\newcommand{\secondres}[1]{{\textcolor{blue}{\textit{#1}}}}
\newcommand{\std}[1]{\scriptsize{$\pm$#1}}

\def\BibTeX{{\rm B\kern-.05em{\sc i\kern-.025em b}\kern-.08em
    T\kern-.1667em\lower.7ex\hbox{E}\kern-.125emX}}
\begin{document}
 
\title{Moving Beyond Functional Connectivity: Time-Series Modeling for fMRI-Based Brain Disorder Classification}

\author{
Guoqi~Yu,
Xiaowei~Hu,~\IEEEmembership{Member,~IEEE},
Angelica~I.~Aviles-Rivero,
Anqi~Qiu,
Shujun~Wang$^{*}$,~\IEEEmembership{Member,~IEEE}%
\thanks{This work was partially supported by the RGC Collaborative Research Fund (No.~C5055-24G), the Start-up Fund of PolyU (No.~P0045999), the Seed Fund of the Research Institute for Smart Ageing (No.~P0050946), the Tsinghua-PolyU Joint Research Initiative Fund (No.~P0056509), the Tsinghua-PolyU Joint Research Initiative Fund (No.~P0056509), and the PolyU UGC funding (No.~P0053716). AIAR gratefully acknowledges the support from YMSC, Tsinghua University.}
\thanks{Guoqi Yu is with the Dept. of Biomedical Engineering, PolyU, Hong Kong, China (e-mail: levi-ack.yu@connect.polyu.hk).}%
\thanks{Xiaowei Hu is with the Sch. of Future Technology, South China University of Technology, China (e-mail: huxiaowei@scut.edu.cn).}%
\thanks{Angelica I. Aviles-Rivero is with the YMSC, Tsinghua University, China (e-mail: aviles-rivero@tsinghua.edu.cn).}%
\thanks{Anqi Qiu is with the Dept. of Health Technology and Informatics, PolyU, Hong Kong, China (e-mail: an-qi.qiu@polyu.edu.hk).}%
\thanks{Shujun Wang is with the Dept. of Biomedical Engineering and the Dept. of Data Science and Artificial Intelligence, PolyU, Hong Kong, China (e-mail: shu-jun.wang@polyu.edu.hk).}%
\thanks{$^{*}$Corresponding author: Shujun Wang.}
}

\maketitle

\begin{abstract}

Functional magnetic resonance imaging (fMRI) enables non-invasive brain disorder classification by capturing blood-oxygen-level-dependent (BOLD) signals. However, most existing methods rely on functional connectivity (FC) via Pearson correlation, which reduces 4D BOLD signals to static 2D matrices—discarding temporal dynamics and capturing only linear inter-regional relationships.
In this work, we benchmark state-of-the-art temporal models (\textit{e.g.}, time-series models: PatchTST, TimesNet, TimeMixer) on raw BOLD signals across five public datasets. Results show these models consistently outperform traditional FC-based approaches, highlighting the value of directly modeling temporal information such as cycle-like oscillatory fluctuations and drift-like slow baseline trends.
Building on this insight, we propose \textbf{DeCI}, a simple yet effective framework that integrates two key principles: (i) \textit{Cycle and Drift Decomposition} to disentangle cycle and drift within each ROI (Region of Interest); and (ii) \textit{Channel-Independence} to model each ROI separately, improving robustness and reducing overfitting.
Extensive experiments demonstrate that DeCI achieves superior classification accuracy and generalization compared to both FC-based and temporal baselines. Our findings advocate for a shift toward end-to-end temporal modeling in fMRI analysis to better capture complex brain dynamics. The code is available at~\url{https://github.com/Levi-Ackman/DeCI}.
\end{abstract}


\section{Introduction}

\IEEEPARstart{F}{unctional} magnetic resonance imaging (fMRI) is widely used for mapping brain activity via blood-oxygen-level-dependent (BOLD) signals~\cite{Worsley2002fmri, Smith2011fmri}, providing key insights into neurological disorders like autism spectrum disorder (ASD), Alzheimer’s disease (AD), and Parkinson’s disease (PD)~\cite{Poldrack2009fmri, xu2024fmri}. However, fMRI data are inherently high-dimensional, noisy, and characterized by complex spatial-temporal dynamics~\cite{birn2006fmrinoise, caballero2017fmrinoise}.
Besides, fMRI dynamics encompass short-term oscillatory fluctuations, long-term baseline variations, and interactions across anatomically defined regions of interest (ROIs), from linear correlations to nonlinear couplings. In this work, we use the term \textbf{Cycle} to denote relatively faster, more oscillatory BOLD fluctuations within the canonical resting-state band (0.01–0.08/0.10 Hz), and \textbf{Drift} to denote relatively slower, baseline-like processes within the same band. These components are influenced by physiological rhythms and hemodynamic processes,\textit{ e.g.}, autonomic and Mayer-wave–related fluctuations around 0.1 Hz, as well as gradual neurovascular and scanner-related changes~\cite{fox2007cycle0,greicius2004cycle1,buckner2008cycle2,smith2009cycle3}. The variability within ROIs and inter-ROI interactions complicate the extraction of reliable patterns, posing challenges for neuroimaging-based classification~\cite{Xu2025ood, chen2022ood, Xu2024transformerfmri}.

Early studies explored simple fully-connected networks (FCNs)~\cite{heinsfeld2018mlpfc}, followed by Graph Convolution Networks (GCNs)~\cite{li2021braingnn} to capture dependencies between ROIs. More recently, Transformer-based models~\cite{kan2022BrainNetTF} have been used to model inter-ROI dependencies, benefiting from self-attention mechanisms.
Most existing methods rely heavily on Static Functional Connectivity (sFC), typically calculated via \emph{Pearson correlation coefficients} between ROI time series~\cite{li2019graph, li2021braingnn, kawahara2017brainnetcnn}. 
While widely used, sFC simplifies the rich 4D fMRI signals into static 2D matrices, discarding critical temporal information within each ROI and capturing only linear relationships~\cite{Linke2020fc, Ciuciu2014fc}. 
Also, sFC assumes stationarity, conflicting with the nonstationary nature of real-world BOLD signals influenced by spontaneous neural activity and transient cognitive states~\cite{huang2020fctemp}, potentially obscuring meaningful connectivity patterns~\cite{cohen2018fctemp}.
Dynamic Functional Connectivity (dFC) methods, such as sliding window correlation and Hidden Markov Models~\cite{rashid2014ydfc, preti2017dyfc}, aim to capture evolving inter-ROI relationships, but still assume stationarity and risk spurious correlations, especially under small sample settings. 

Both sFC and dFC depend on expert-chosen parameters and overlook intra-ROI dynamics like physiological rhythms and baseline drift.
Motivated by these limitations, we ask: \emph{\textbf{Can we surpass FC by directly learning intra-ROI temporal patterns and inter-ROI dependencies from raw BOLD signals?}}

\begin{figure}[t]
  \centering
  \includegraphics[width=0.48\textwidth]{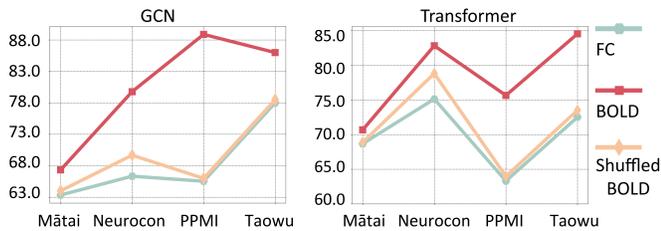}
\caption{For each identical backbone (GCN; Transformer), we trained three instances that differ only in input: (i) FC, (ii) raw BOLD signals, and (iii) Shuffled-BOLD that applies the same temporal permutation to every ROI, thus preserving FC while ablating temporal order. The BOLD models outperform FC; Shuffled-BOLD degrades to near-FC, isolating the contribution of temporal dynamics.}
  \label{fig:fmri_ts}
  \vspace{-4mm}
\end{figure}

To address this question, we compare GCN and Transformer on four fMRI datasets, as both effectively capture FC-based interactions among ROIs~\cite{ktena2018fcgnn, hayat2024fctransformer} and can also model temporal and inter-channel dependencies in multivariate time series~\cite{Wangtslb2024}. 
Our goal is to assess whether these deep models can serve as end-to-end alternatives to traditional FC-based pipelines for brain dynamics modeling. 
As shown in Fig.~\ref{fig:fmri_ts}, models trained directly on raw BOLD signals outperform those using Pearson-based FC features. 
When BOLD timepoints are shuffled, preserving FC but disrupting temporal dynamics, performance drops, yet still remains comparable to that of FC-based models.
These results indicate that (i) BOLD signals offer discriminative temporal patterns beyond static FC, which deep models can exploit, and (ii) much of FC’s discriminative power is also accessible from BOLD signals alone.

Motivated by these insights, we benchmark several cutting-edge temporal (time series) models,
including PatchTST~\cite{Nie2022patchtst}, TimesNet~\cite{wu2022timesnet}, and TimeMixer~\cite{wang2023timemixer} for fMRI classification. 
Results across six public fMRI datasets confirm that such general time series analysis purpose models significantly outperform traditional FC-based approaches. 
However, these temporal models still face generalization challenges due to two core issues:
(i) the entanglement of various temporal components within ROI signals (such as high-frequency cycles, low-frequency drift, and gradual changes from factors like head motion), and (ii) the tendency of existing deep models to overfit to spurious inter-ROI dependencies, especially in low-data settings.

To address these challenges, we develop DeCI (Decomposition and Channel-Independence), a novel end-to-end framework that integrates two principles from general time series modeling:
(i) \textbf{Cycle and Drift Decomposition}, which separates the band-limited BOLD signals into relatively faster, oscillatory components (``Cycle") and slower, baseline-like components (``Drift") within the canonical 0.01–0.08/0.10 Hz resting-state band, helping disentangle signal characteristics; and
(ii) \textbf{Channel-Independence}, which processes each ROI independently using a shared model before aggregating inter-regional information, thereby mitigating overfitting.
By first extracting clean intra-ROIs dynamic features and only later modeling inter-ROIs dependencies, DeCI significantly improves robustness and generalization in fMRI classification.

Our contributions can be summarized as follows.
\begin{itemize}
\item We benchmark traditional FC against state-of-the-art time-series models (\textit{e.g.}, PatchTST, TimesNet), showing that automatic feature learning greatly outperforms handcrafted methods.

\item We formulate DeCI, a novel deep learning model for fMRI classification that integrates Cycle and Drift Decomposition with Channel-Independence to better capture both intra-ROIs and inter-ROIs features.

\item Experimental results on six public fMRI datasets demonstrate that DeCI consistently outperforms competing methods by a significant margin.
\end{itemize}

The remaining of this paper is organized as follows.
We review the related work in Section II, present the benchmarking in Section III, and elaborate the Temporal decomposition, Channel-Independence, and proposed DeCI in Section IV. The experiment details and results are presented in Section V. We further discuss our method in Section VI and draw the conclusions in Section VII.
\section{Related Work}
In this section, we first review relevant literature on fMRI-based brain network classification, followed by recent advances in deep learning for general time series analysis.

\subsubsection{fMRI-based brain network classification} 
The classification of fMRI data has evolved significantly in recent years, transitioning from traditional handcrafted biomarker extraction~\cite{bullmore2009Biomarkerfmri} to machine learning-based approaches~\cite{richiardi2011mlfmri, Elia2008mlfmri, KHOSLA2019mlfmri}. Recently, deep learning techniques, such as Graph Neural Networks (GNNs)~\cite{Gadgil2020gnnfmri, parisot2018gnnfmri,Yao2021gnnfc}, have emerged as powerful tools for fMRI classification, enabling the modeling of complex brain networks and spatial dependencies across ROIs. Additionally, Transformer-based models~\cite{Wang2024transformerfc,Han2025transformerfc,Shehzad2025Transformerfc,yu2024Transformerfc} have been applied to fMRI data, benefiting from their self-attention mechanism to capture dependencies across brain regions. 

Despite achieving impressive performance, these models rely predominantly on Static Functional Connectivity (sFC) features calculated using Pearson correlation. 
However, functional connectivity primarily captures linear dependencies between brain regions, fails to reflect more complex inter-ROIs dependencies, and completely neglects intra-ROIs dynamics. While Dynamic Functional Connectivity (dFC)~\cite{preti2017dyfc,rashid2014ydfc} solves these problems to a certain degree, it still relies on handcrafted features or windowing strategies with strong predefined assumptions, which restrain its flexibility and capacity. 

\subsubsection{Advancements in general time series analysis} 
Recent general time series analysis has witnessed groundbreaking developments by exploring the potential of various architectures, including Transformer~\cite{Zhou2020informer,Liu2021pyraformer,zhou2022fedformer}, Mamba~\cite{wang2024mamba,patro2024simba,Zhang2024mamba}, Multi-layer Perceptron (MLPs)~\cite{Das2023TiDE,lin2024cyclenet,lin2024sparsetsf}, Temporal Convolution Networks (TCNs)~\cite{wang2023micn,Liu2021scinet}, and even Large Language Models (LLMs)~\cite{jin2023timellm,liu2024timellm}. Beyond architectural innovations, the effective adoption of two strategies has significantly advanced the development of the field, i.e., \emph{(i) Seasonal and Trend Decomposition (or Cycle and Drift Decomposition in this work)}~\cite{Wu2021autoformer,wang2023timemixer,yu2024leddam} and \emph{(ii) Channel-Independence (CI) strategy}~\cite{Nie2022patchtst,zeng2023dlinear}. \textbf{Seasonal and Trend Decomposition} disentangle meaningful patterns from raw time series (such as BOLD signals) into short-term fluctuations (high-frequency cycle) and long-term variations (low-frequency drift) components, delivering more precise temporal variations modeling~\cite{Zhang2022tdformer}. 
\textbf{Channel-Independence} models each channel (such as ROI) separately using a shared backbone, without considering other channels~\cite{Nie2022patchtst}. It assumes that all the time series in the set come from the same process and fit a uniform univariate function~\cite{han2023ci1, chen2024ci3}. Despite its simplicity, this strategy yielded surprisingly strong performance and robustness. 

While these strategies have achieved remarkable success in general time series applications, they remain underexplored in the context of fMRI classification.
Besides, directly applying temporal models to fMRI is \textit{no free lunch}: without fMRI-aware inductive biases, models tend to mix ROIs spuriously, and fail to decouple baseline drift and physiological cycles. CycleNet focuses solely on cycle components while neglecting drift; Leddam conducts a shallow, one-step cycle and drift split that is misaligned with the multi-scale nature of fMRI signals~\cite{Smith2011fmri,birn2006fmrinoise}. PatchTST applies CI modeling at the patch-level and then fuses representations of all channels before prediction, which can reintroduce spurious inter-channel/ROI interaction. 
So in this work, we build the first fMRI classification benchmark grounded in modern temporal models to explore their possibility. Then, we propose \textbf{DeCI}, which employs \emph{deep Cycle and Drift decomposition} via progressive residual extraction and CI modeling at the \emph{ROI (series) level} with fusing \emph{at the logit level}. Our DeCI can handle both the deep entanglement of cycle and drift and the spurious inter-ROI correlations.

\section{Benchmark}

To assess the benefit of end-to-end temporal modeling over traditional FC-based methods, we establish a comprehensive benchmark that \textbf{\textit{(i)}} evaluates state-of-the-art time series methods on raw BOLD signals and \textbf{\textit{(ii)}} compares them against classical FC-based approaches. 

\textbf{Construction. \quad} We preprocess six public fMRI datasets (Mātai~\cite{xu2023maitai}, TaoWu~\cite{badea2017taowu}, Neurocon~\cite{badea2017taowu}, PPMI~\cite{badea2017taowu}, ABIDE~\cite{craddock2013abide}, and ADNI~\cite{aisen2010adni0,aisen2015adni1,weiner2017adni2}, detailed in Table~\ref{tab:dataset}) using the AAL (Automated Anatomical Labeling) atlas~\cite{tzourio2002aal116} to obtain \textbf{116} ROIs per subject. We then evaluate seven modern time series models—including Transformers (Leddam~\cite{yu2024leddam}, iTransformer~\cite{LiuiTransformer}, PatchTST~\cite{Nie2022patchtst}), Temporal Convolutional Networks (TimesNet~\cite{wu2022timesnet}, ModernTCN~\cite{donghao2024moderntcn}), and Multi-layer Perceptron (TimeMixer~\cite{wang2023timemixer}, TSMixer~\cite{chen2023tsmixer})—alongside four classical FC-based classifiers (Deep learning: BrainNetTF~\cite{kan2022BrainNetTF}, BrainOOD~\cite{Xu2025ood}; Mechine learning: SVM~\cite{Elia2008mlfmri}, Random Forest~\cite{KHOSLA2019mlfmri}). We also evaluated six cutting-edge fMRI classifiers, including attention-based classifiers (STAGIN~\cite{kim2021attn3}, PSCRAttn~\cite{yang2021attn4}), multi-view models (MVHO~\cite{zhao2021multiview1}, SimMVF~\cite{li2025multiview2}), and dynamic FC-based fMRI classifiers (MDGL~\cite{ma2023mdgldfc1}, BrainTGL~\cite{liu2023braintgldfc2}). We represent the detailed results in Table~\ref{tab:0bench_std}\&\ref{tab:1bench_std}\&\ref{tab:2bench_std}. All methods are trained and tested under identical five-fold cross-validation protocols, with five runs and five metrics: Accuracy, Precision (macro-averaged), Recall (macro-averaged), F1-Score (macro-averaged), and AUROC (Area Under the Receiver Operating Characteristic curve, macro-averaged).

\textbf{Results. \quad} 
We develop \textbf{\textit{the first benchmark for fMRI classification using state-of-the-art time series models}}. These models consistently outperform classical FC-based classifiers across all datasets (Tables~\ref{tab:0bench_std}\&\ref{tab:1bench_std}\&\ref{tab:2bench_std}). Averaged over five metrics, the best temporal model achieves up to a \textbf{41.45\%} improvement on the PPMI dataset, with a mean gain of \textbf{4.39\%} (After removing the highest 41.45\% and lowest 0.6\%). Among these temporal models, the Transformer-based models generally deliver the highest performance, but the MLP-based methods achieve comparable performance with relatively lower resource consumption. The leading temporal model is \textbf{Leddam}~\cite{yu2024leddam}, which is a Transformer-based variant and based on Cycle and Drift Decomposition. 
These findings highlight the effectiveness of direct temporal modeling in capturing intra-ROI and inter-ROI patterns, and support a paradigm shift from FC-based and dFC-based pipelines to end-to-end temporal learning for fMRI decoding.

\section{Methodology}
\begin{figure*}[th]
  \centering
  \includegraphics[width=0.99\textwidth]{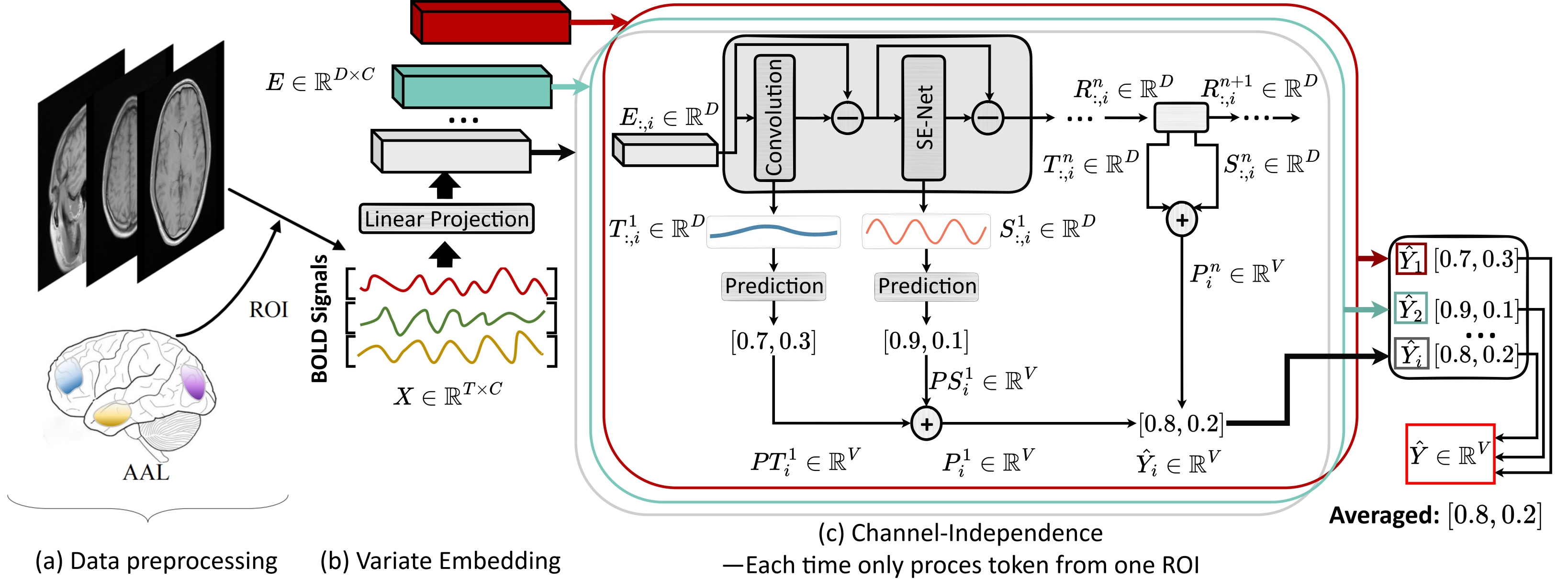}
  \caption{General structure of \textbf{DeCI}, a Channel-Independent framework. We use Linear Convolution networks to extract the $Drift$ pattern, and Nonlinear Squeeze and Excitation Networks (SE-Net) to model the $Cycle$ pattern. The final classification result is the sum of multiple logits generated using $Drift$ and $Cycle$ information extracted from each DeCI Block.}
  \label{fig:deci}
\end{figure*}
This section first introduces the preliminaries of the Cycle and Drift Decomposition (\textit{i.e.}, Seasonal and Trend Decomposition in general time series analysis) and Channel-Independence. Then, present our framework, \textbf{DeCI}, which integrates Cycle and Drift Decomposition and Channel-Independent modeling for robust fMRI classification.

\subsection{Preliminaries}

Consider an input fMRI-based BOLD signal sample $X \in \mathbb{R}^{T \times C}$, obtained after standard preprocessing and band-pass filtering in the canonical 0.01–0.08/0.10 Hz resting-state band, and its corresponding label $Y \in \mathbb{R}^{V}$, where $T$ denotes the number of timestamps (length of the BOLD signals), $C$ represents the number of channels (ROIs), and $V$ denotes the number of classification categories (\textit{e.g.}, disease classes). 
Traditional classification methods typically rely on functional connectivity.           
They first compute a Pearson correlation matrix $X_{fc} \in \mathbb{R}^{C \times C}$ from $X$, which captures pairwise linear dependencies between ROIs, and then use $X_{fc}$ as the model input to predict $Y$. 
In contrast, general time series models bypass handcrafted connectivity features and directly use the raw BOLD signal $X \in \mathbb{R}^{T \times C}$ as input, allowing the model to automatically learn both temporal dynamics within each ROI and inter-ROI dependencies. 

But the deployment of general time series models needs to tackle two generalization challenges: (i) the entanglement of multiple temporal components (e.g., high-frequency oscillatory dynamics and low-frequency baseline trends within the resting-state band) within ROI signals, and (ii) the spurious inter-ROIs dependencies caused by noise and small sample size. So we borrow two principles from general time series modeling:
(i) \textbf{Cycle and Drift Decomposition}, which separates BOLD signals into short-term fluctuations (high-frequency cycle) and long-term variations (low-frequency drift), helping disentangle intra-ROIs signal characteristics; and (ii) \textbf{Channel-Independence}, which processes each ROI independently using a shared model before aggregating inter-ROIs information.

\subsubsection{Definition 3.1 Cycle and Drift Decomposition}
In general time series analysis, given a univariate time series $x \in \mathbb{R}^{T}$, the Seasonal and Trend Decomposition (STD) assumes it is the additive combination of a trend $\mu \in \mathbb{R}^{T}$, and a seasonal $\omega \in \mathbb{R}^{T}$: $x =\mu+\omega.$ We borrow this principle and denote it as \textbf{\textit{Cycle and Drift Decomposition}} for better understanding in the context of fMRI-based BOLD signal modeling.

Seasonal-Trend Decomposition (or Cycle and Drift Decomposition) disentangles a signal into Seasonal and Trend, the former representing the periodic short-term fluctuations (the relatively higher-frequency cycle), the latter reflecting the smooth long-term variations (the relatively lower-frequency shifts), within the resting-state band, which can significantly improve pattern interpretability and discriminability~\cite{fawaz2020stdclass}. 

\subsubsection{Definition 3.2 Channel-Independence}
Given a $C$-channel multivariate time series $X \in \mathbb{R}^{T \times C}$, the Channel-Independent strategy assumes that all the time series in the set come from the same process and fit a uniform univariate function~\cite{han2023ci1}. Thus, each channel (ROI) is modeled separately using a shared function $ f_{CI}: \mathbb{R}^T \rightarrow \mathbb{R}^V$, without considering information from other channels.

The CI model, which processes each channel independently, shows strong robustness to noise and nonstationarity~\cite{Nie2022patchtst,han2023ci1}, which are common issues in BOLD signals.

\subsection{The DeCI Framework}
In this section, we formulate a novel temporal model:
\textbf{DeCI}, depicted in Fig.~\ref{fig:deci}. Initially, we tokenize the BOLD signal input $X \in \mathbb{R}^{T \times C}$, into \textbf{Variate Embedding} $ E \in \mathbb{R}^{D  \times C}$~\cite{LiuiTransformer}, where the whole series of a ROI is tokenized into a token using a linear projection. This makes sure the unique patterns of each ROI are well-preserved. Then, $E$ goes through $N$ \textbf{DeCI Block} in a Channel-Independent way, where the tokens of all channels (ROI) are processed individually using a shared model. Since no explicit interaction between these ROIs is performed, such a strategy can suppress noise propagation between representations of different ROIs. Each Block generates two classification logits based on extracted $Drift$ and $Cycle$ information, respectively. We follow convention from previous time series works~\cite{Zhang2022tdformer,wang2023micn,yu2024leddam}, where $Drift$ is modeled using Linear models, a convolution neural network ($\operatorname{Convolution}$ in Fig.~\ref{fig:deci})~\cite{yu2024leddam}, and $Cycle$ using Nonlinear models, a Squeeze and Excitation networks~\cite{hu2018senet} ($\operatorname{SE-Net}$ in Fig.~\ref{fig:deci}). \textbf{Drift/trend} is a smooth, low-frequency component, whereas \textbf{cycle/seasonal} is oscillatory/high-frequency patterns. A 1D convolution is a finite-impulse-response linear time-invariant filter that naturally implements \textbf{low-pass smoothing}, the standard trend estimator in classical time series analysis and modern TS backbones~\cite{cleveland1990stl,stock1998lino,dama2021analysis,Wu2021autoformer}. By contrast, cycle components in fMRI are often \textbf{nonlinear}, exhibiting time-/ROI-varying amplitude and phase; fixed linear filters are therefore insufficient. A \textbf{Squeeze-and-Excitation Network} ($\operatorname{SE-Net}$) branch provides nonlinear, input-conditioned multiplicative gating (adaptive gain control), enabling selective amplification/attenuation of oscillatory features and can effectively capture high-frequency patterns~\cite{box2015high,harvey1990cycle,durbin2012cycle,hu2018senet}. Together, linear Conv1D for Drift and nonlinear SE-Net for Cycle, yield a theory-grounded inductive bias that is robust to BOLD non-stationarities.
The extraction of each pattern is based on the \textbf{\textit{Residual}} of the previous representation~\cite{Oreshkin2020nbeats}. 

\subsubsection{Progressive Cycle and Drift Decomposition} 
Given the input token of $n$-th Block from $i$-th ROI (channel) $R^n_{:,i}\in \mathbb{R}^{D},\quad i=1,\dots,N$ ($R^{1}_{:,i}=E_{:,i}$), we first get its Drift pattern $T^n_{:,i}\in \mathbb{R}^{D}$ and Drift prediction $PT^n_{i}\in \mathbb{R}^{V}$ by $\operatorname{Convolution}$ defined as below:
\begin{align}
\tilde{R}^n_{:,i} &= \text{Padding}(\mathbf{0}, R^n_{:,i}),  \quad R^n_{:,i} \in \mathbb{R}^{D},\quad\tilde{R}^n_{:,i}\in {\mathbb{R}^{D+K-1}}, \notag\\
T^n_{:,i} &= \text{Dropout}\left( \text{Conv1D}(\tilde{R}^n_{:,i}; W)) \right), \quad W \in \mathbb{R}^{K}, \notag\\
PT^n_{i} =& T^n_{:,i}W_{t}+b_{t}, \quad T^n_{:,i} \in \mathbb{R}^{D},  \quad W_{t} \in \mathbb{R}^{D\times V}, \quad b_{t} \in \mathbb{R}^{V}.
\end{align}
This is a classical 1D convolution with kernel size $K$, stride $J=1$, no dilation, and $K-1$ zero padding in the front to ensure the same input-output dimension. We set the kernel size equal to the model dimension, i.e., $K=D$, to ensure a full receptive field. Then, the residual $L^n_{:,i}=R^n_{:,i}-T^n_{:,i}$ is used to extract Cycle pattern $S^n_{:,i}\in \mathbb{R}^{D}$ and Cycle prediction $PS^n_{i}\in \mathbb{R}^{V}$ by $\operatorname{SE-Net}$ defined as below:
\begin{align}
&G = \text{Sigmoid} \left( ( \text{GELU}(\text{Dropout}(L^n_{:,i}W_1 + b_1)) ) W_2 + b_2 \right) \notag\\
&H = (L^n_{:,i} \odot G) W_g + b_g, \quad Q_1 = \text{LayerNorm}(H + L^n_{:,i}), \notag\\
&Q_2 = \text{GELU}(\text{Dropout}(Q_1 W_3 + b_3)) W_4 + b_4,\notag\\
&S^n_{:,i} = \text{LayerNorm}(Q_2 + Q_1), \notag\\
&W_g, W_1, W_2,W_3,W_4, \in \mathbb{R}^{D\times D}, \quad  b_g, b_1, b_2,b_3,b_4, \in \mathbb{R}^{D}, \notag\\
&PS^n_{i} = S^n_{:,i}W_{s}+b_{s},  \quad W_{s} \in \mathbb{R}^{D\times V}, \quad b_{s} \in \mathbb{R}^{V}.
\end{align}
This $\operatorname{SE-Net}$ branch provides nonlinear, input-conditioned multiplicative gating (adaptive gain control), enabling selective amplification/attenuation of oscillatory features and effectively modeling the high-frequency cycle patterns.

So, the overall prediction generated in $n$-th Block using representation from $i$-th ROI (channel) is $P^n_i=\frac{1}{2}(PT^n_{i}+PS^n_{i})$. The residual $R^{n+1}_{:,i}=L^n_{:,i}-S^n_{:,i}=R^{n}_{:,i}-T^n_{:,i}-S^n_{:,i}$ is passed to the next Block. By progressively decomposing Cycle and Drift components across multiple depths and aggregating logits from each block, this deep ensemble approach captures multi‑scale disentangled patterns.

\subsubsection{Channel-Independent design} 
Our DeCI model adopts a channel-independent strategy~\cite{Nie2022patchtst, han2023ci1}, where, for each input BOLD signal with $C$ ROIs (channels), only one channel is processed at a time. Based on the progressive Cycle and Drift Decomposition framework mentioned above, a sequence of DeCI blocks is applied to generate a series of individual classification logits $P^n_i,\quad i=1,\dots, C, \quad n=1,\dots, N$ for the $i$-th channel at $n$-th Block. The final prediction is then obtained by averaging over all these logits:
\begin{align}
\hat{Y}=\frac{1}{C}\frac{1}{N}\sum_{i=1}^{C}\sum_{n=1}^{N} P^n_i=\frac{1}{C}\frac{1}{2N}\sum_{i=1}^{C}\sum_{n=1}^{N} (PT^n_i+PS^n_i).
\end{align}
Since the final result is the sum of all logits from all blocks, this forms a \textbf{Deep Ensemble Learning}~\cite{kearns1994ensemble,suk2017ensemble} framework, hereby empowering the model with enhanced resilience against overfitting. 
\begin{table}[tp]
\small
   \caption{Statistics of the datasets. Each subject has a graph (brain connectivity network) generated using the AAL Parcellation Method, resulting in \textbf{116} ROIs, or 116 \textbf{Channels} ($C=116$). The length of the BOLD signals corresponds to the number of node features, $ T$.}
    \vspace{-2mm}
    \label{tab:dataset}
    \centering
    \resizebox{0.5\textwidth}{!}{
    \scalebox{1}{\begin{tabular}{ccccc}
    \toprule
       \textbf{Dataset} & \textbf{Condition} & \textbf{Subjects} & \textbf{Classes} & \textbf{Signals Length}\\
    \midrule
        Mātai~\cite{xu2023maitai}    & mTBI & 60   & 2   & 200\\
        TaoWu~\cite{badea2017taowu}    & Parkinson  & 40   & 2   & 239\\
        Neurocon~\cite{badea2017taowu} & Parkinson  & 41   & 2   & 137\\
        PPMI~\cite{badea2017taowu}     & Parkinson  & 209  & 4   & 210 \\
        ADNI~\cite{weiner2017adni2}   & Alzheimer’s Disease &720 &3 &197\\
        ABIDE-300~\cite{craddock2013abide}    & Autism     & 138 & 2   & 300 \\
        ABIDE-240~\cite{craddock2013abide}    & Autism     & 120 & 2   & 240 \\
        ABIDE-180~\cite{craddock2013abide}    & Autism     & 236 & 2   & 180 \\
        ABIDE-120~\cite{craddock2013abide}    & Autism     & 134 & 2   & 120 \\
    \bottomrule
    \end{tabular}}
    }
    \vspace{-2mm}
\end{table}

\section{Experimental Results on Benchmark Datasets}

\subsection{Benchmark Configuration}
\emph{Notably, this is the first study to establish a benchmark for fMRI classification using cutting-edge temporal models.}

\subsubsection{Datasets}
The datasets used for evaluation are well-processed brain network datasets based on fMRI data, processed using the AAL (Automated Anatomical Labeling)~\cite{tzourio2002aal116} protocol, resulting in $116$ ROIs. A summary of the datasets is in Table~\ref{tab:dataset}. 
They are:
(i) \textbf{Mātai}~\cite{xu2023maitai}: A longitudinal study designed to detect subtle brain changes due to a season of contact sports.
(ii) \textbf{TaoWu} and \textbf{Neurocon}~\cite{badea2017taowu}: Early image datasets for Parkinson’s disease (PD).
(iii) \textbf{PPMI}~\cite{badea2017taowu}: Aimed at identifying biological markers for PD, onset, and progression.
(iv) \textbf{ADNI}~\cite{weiner2017adni2}: A longitudinal multisite study for the early detection and tracking of Alzheimer’s Disease (AD).
(v) \textbf{ABIDE}~\cite{craddock2013abide}: A global initiative aggregating functional brain imaging data for Autism Spectrum Disorder (ASD).

\subsubsection{Data Preprocessing Details}
All datasets and preprocessing strictly \textbf{follow the benchmark pipeline of \textit{Xu et al.}~\cite{xu2023maitai}.} 

\textbf{Motion correction \& censoring.} Rigid-body realignment; motion regressors expanded to 24-parameter model (translations/rotations + derivatives + squares). Volumes exceeding conventional motion thresholds are censored (framewise displacement and DVARS), following standard practice~\cite{power2012spurious}.

\textbf{Nuisance regression/denoising.} Regressors include 24p motion, mean WM/CSF (and aCompCor components when applicable), linear/quadratic trends; optional ICA-AROMA when provided by the benchmark, consistent with widely used denoising schemes~\cite{behzadi2007compcor}.

\textbf{Band-pass filtering.} Temporal filtering in the low-frequency resting-state band (e.g., 0.01–0.08/0.10 Hz) using zero-phase filtering; filtering is ordered with nuisance regression to avoid reintroducing noise~\cite{hallquist2013nuisance}. DeCI is subsequently applied to these band-pass–filtered ROI time series and thus decomposes BOLD dynamics within this frequency band.

\textbf{Parcellation \& time series extraction.} ROI time series are extracted after standard spaces/harmonization using AAL (116 ROIs) atlases~\cite{tzourio2002aal116}.

\textbf{Temporal harmonization \& scaling.} For datasets with variable length, we retain the largest subset with identical length (e.g., ADNI $T{=}197$) or split them into multiple subsets based on their length (e.g., ABIDE-120/180/240/300). Then, we z-score the ROI series within the session. 

All preprocessing parameters and seeds are provided in the code repository for exact traceability.

\subsubsection{Evaluation Models} 
The baseline evaluation models, consisting of \textit{7} state-of-the-art general time series analysis methods, \textit{4} classical FC-based fMRI data classification methods, and \textit{6} cutting-edge fMRI classifiers, are categorized as follows:

\textbf{\textit{(1) Cutting-edge general time series analysis methods:}} 
(i) \textbf{Transformer-based}: Leddam~\cite{yu2024leddam}, iTransformer~\cite{LiuiTransformer}, PatchTST~\cite{Nie2022patchtst}
(ii) \textbf{Temporal Convolution Networks (TCN)}: TimesNet~\cite{wu2022timesnet}, ModernTCN~\cite{donghao2024moderntcn}
(iii) \textbf{Multi-Layer Perceptron (MLP)}: TimeMixer~\cite{wang2023timemixer}, TSMixer~\cite{chen2023tsmixer}.

\textbf{\textit{(2) Classical FC-based fMRI data classification methods:}} 
(i) Deep learning: BrainNetTF~\cite{kan2022BrainNetTF}, BrainOOD~\cite{Xu2025ood}
(ii) Machine learning: SVM~\cite{Elia2008mlfmri}, Random Forest~\cite{KHOSLA2019mlfmri}.

\textbf{\textit{(3) Various cutting-edge fMRI data classification methods:}} 
(i) attention-based methods: STAGIN~\cite{kim2021attn3}, PSCRAttn~\cite{yang2021attn4}
(ii) multi-view models: MVHO~\cite{zhao2021multiview1}, SimMVF~\cite{li2025multiview2}
(iii) dFC-based classifiers: MDGL~\cite{ma2023mdgldfc1}, BrainTGL~\cite{liu2023braintgldfc2}.

\subsubsection{Implementation Details}
We evaluate the models using \textbf{five-fold cross-validation}, with results averaged from five runs. All methods follow the same hyperparameter search protocol, which includes the number of layers-$N\in\{1,2\}$, model dimension-$D\in\{64,128\}$, and dropout ratio-$dp\in\{0.0,0.2\}$. The final selection of each hyperparameter for DeCI on each dataset is listed in Table~\ref{tab:hyper_final}. The performance of each model is measured using the following five metrics: Accuracy, Precision (macro-averaged), Recall (macro-averaged), F1-Score (macro-averaged), and AUROC (Area Under the Receiver Operating Characteristic curve,macro-averaged). 
\textit{Implementation codes for the baselines, the DeCI model, and hyperparameter settings are available at \url{https://github.com/Levi-Ackman/DeCI}.}
\begin{table}[t]
\centering
\small
\caption{Final selected hyperparameters for DeCI.}
\label{tab:hyper_final}
\resizebox{0.3\textwidth}{!}{%
\begin{tabular}{lccc}
\toprule
\textbf{Dataset} & \boldmath{$N$} & \boldmath{$D$} & \boldmath{$dp$} \\
\midrule
Mātai            & 2 & 64 & 0.0 \\
TaoWu            & 2 & 64 & 0.2 \\
Neurocon         & 2 & 64 & 0.0 \\
PPMI             & 1 & 64 & 0.0 \\
ADNI             & 1 & 64 & 0.0 \\
ABIDE-120        & 1 & 64 & 0.2 \\
ABIDE-180        & 2 & 64 & 0.2 \\
ABIDE-240        & 2 & 64 & 0.0 \\
ABIDE-300        & 2 & 64 & 0.0 \\
\bottomrule
\end{tabular}
}
\end{table}

\begin{figure}[t]
  \centering
  \includegraphics[width=0.48\textwidth]{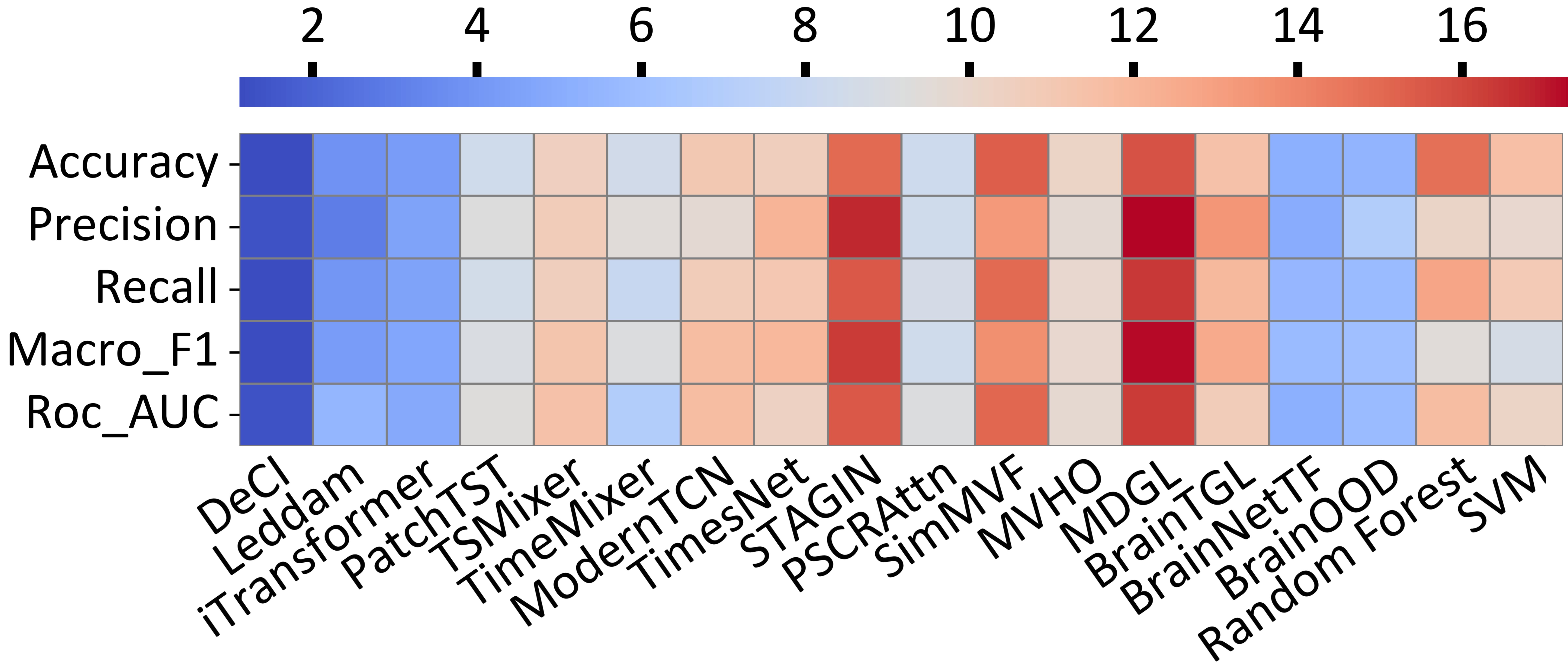}
  \caption{Average rank of all benchmarks across five metrics.}
  \label{fig:rank}
\end{figure}

\begin{figure}[t]
  \centering
  \includegraphics[width=0.48\textwidth]{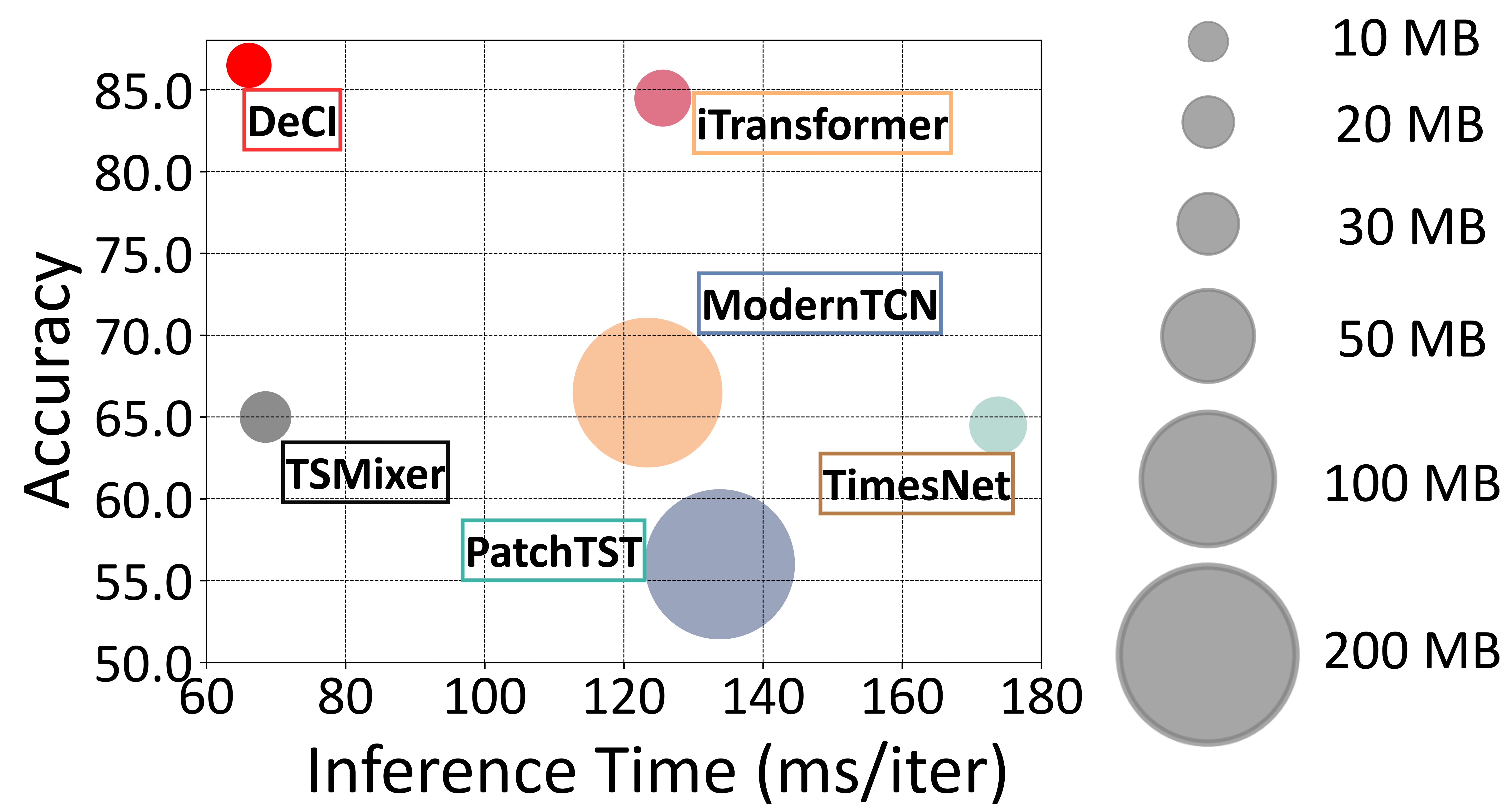}
  \caption{Model efficiency comparison on the TaoWu, comparing Accuracy, inference time, and memory footprint.}
  \label{fig:efficiency}
\end{figure}

\subsection{Comprehensive Results on Benchmark Datasets}
\subsubsection{Main results}
Table~\ref{tab:0bench_std}\&\ref{tab:1bench_std}\&\ref{tab:2bench_std} present benchmark results comparing 17 existing models with our new proposed DeCI, which highlight the significant performance gap between traditional Pearson correlation methods (row 1-4), cutting-edge fMRI classifiers (row 5-10), and recent state-of-the-art time-series methods (row 11-18) for fMRI-based brain network classification. 
The results demonstrate that general time series analysis methods outperform the classical FC-based fMRI data classification method and cutting-edge fMRI classifiers.
This validates the effectiveness of modern time series models in automatically learning complex features from fMRI data, offering a substantial improvement over classical FC/dFC-based techniques in terms of classification accuracy and robustness across multiple datasets.

\subsubsection{Quantitative results}

The proposed DeCI model demonstrates consistent top-tier performance, excelling in all metrics on most datasets, and remains second on the remaining one (PPMI). 
Overall, DeCI achieves 7/9 top-1 Precision and AUROC, and 8/9 top-1 rankings across the other three metrics, reflecting its remarkable performance. Fig.~\ref{fig:rank} provides an overview of the average rank across six datasets and five metrics, with DeCI showing a dominant superiority over other methods. Interestingly, Leddam ranks second overall; like DeCI, it also uses Cycle and Drift Decomposition, reinforcing the idea that Cycle and Drift Decomposition provide a powerful inductive bias for precise and stable classification.

\subsubsection{Efficiency analysis}

DeCI exhibits \textit{{linear}} complexity concerning the number of channels (ROIs), compared to the quadratic complexity of Transformer-based methods. As shown in Fig.~\ref{fig:efficiency}, DeCI achieves the highest classification accuracy while maintaining substantially lower memory usage and faster inference.

\subsubsection{Robustness to Parcellation Atlas} 
We additionally evaluated the Schaefer atlas (100 ROIs)~\cite{schaefer2018localglobal}, Ward atlas (100 ROIs)~\cite{Thirion2014ward100}, and Harvard atlas (48 ROIs)~\cite{HarvardOxford2012havard48} to assess the robustness of our DeCI and some representative models to the parcellation atlas. As reported in Table~\ref{tab:atlas}, DeCI remains SOTA across all atlases, and the performance fluctuations across parcellations are minor, indicating its strong robustness. 

\subsubsection{Statistical Significance Analysis (DeCI vs.\ Leddam)}
Paired two-sided $t$-tests on the 5-fold results (Table~\ref{tab:significance}) show that \textbf{DeCI} significantly outperforms \textbf{Leddam}. On \textbf{ADNI}, DeCI improves Accuracy by $+11.25$\,pp \([8.24, 14.28]\), $d_z=4.63$, and F1-Score by $+22.13$\,pp \([13.92, 30.33]\), $d_z=3.35$ (both $p<0.001$). On \textbf{Mātai}, Accuracy and F1-Score improve by $+3.34$\,pp \([0.35, 6.31]\), $d_z=1.39$, and $+4.04$\,pp \([0.32, 7.76]\), $d_z=1.35$ (both $p<0.05$). On \textbf{Neurocon}, DeCI yields $+4.11$\,pp in Accuracy \([0.75, 7.47]\), $d_z=1.81$, and $+2.66$\,pp in F1-Score \([0.18, 5.14]\), $d_z=1.33$ (both $p<0.05$). On \textbf{TaoWu}, DeCI yields $+11.50$\,pp in Accuracy \([7.96, 15.04]\), $d_z=4.03$, and $+13.83$\,pp in F1-Score \([9.71, 17.93]\), $d_z=4.17$ (both $p<0.001$). With $n{=}5$ folds, $d_z>1.2$ denotes strong, stable improvements, met on Neurocon and Mātai and vastly exceeded on ADNI and TaoWu, highlighting DeCI’s effectiveness and robustness against overfitting on small sample datasets (e.g., Neurocon and TaoWu).

\begin{table*}[!h]
\small
\centering
\caption{Performance (mean$_{std}$) of all methods on Five-fold cross-validation across Mātai, Neurocon, and PPMI. All results are obtained over five runs. The ``Avg'' metric means the average value of the five evaluation metrics for convenient comparison. The best performance is highlighted in \boldres{red}, and the second is highlighted in \secondres{blue}.}
\begin{threeparttable}
\resizebox{0.86\textwidth}{!}{%
\begin{tabular}{c|rcccccc}
\midrule
\multicolumn{1}{c}{} &Metrics & Accuracy & Precision & Recall & F1-Score & AUROC & Avg\\ 
\midrule
\multirow{18}{*}{\rotatebox{90}{\textbf{Mātai}}}
&\lineorange{SVM} &62.00\std{2.21} &57.94\std{7.31} &58.29\std{2.18} &54.71\std{3.05} &57.03\std{9.11} &57.99\std{4.77} \\
&\lineorange{Random Forest} &50.67\std{4.55} &47.76\std{5.46} &48.91\std{4.93} &47.29\std{5.28} &51.71\std{7.48} &49.27\std{5.54} \\
&\lineorange{BrainOOD} &63.33\std{4.47} &45.97\std{9.34} &56.91\std{6.42} &48.17\std{9.44} &53.71\std{5.97} &53.62\std{7.13} \\
&\lineorange{BrainNetTF} &68.67\std{1.94} &70.64\std{7.56} &63.89\std{2.32} &60.13\std{3.77} &63.31\std{3.71} &65.33\std{3.86} \\
&\linegreen{BrainTGL}  &60.33\std{3.23} &36.46\std{9.22} &53.20\std{5.21} &41.95\std{7.91} &55.20\std{8.93} &49.43\std{7.30}\\
&\linegreen{MDGL}      &58.67\std{0.67} &31.11\std{3.89}  &50.86\std{1.71} &38.00\std{2.31} &50.63\std{6.09} &45.85\std{2.93}\\
&\linegreen{MVHO}      &64.33\std{2.71} &55.49\std{9.23} &59.49\std{4.30} &51.65\std{6.47} &59.60\std{4.41} &58.11\std{6.22}\\
&\linegreen{SimMVF}    &52.67\std{5.12} &34.28\std{7.13}  &47.31\std{1.93} &38.72\std{3.41} &47.31\std{1.93} &44.06\std{3.90}\\
&\linegreen{PSCRAttn}  &66.00\std{2.91} &60.34\std{8.41} &60.69\std{4.06} &54.76\std{6.22} &63.49\std{7.96} &61.06\std{6.31}\\
&\linegreen{STAGIN}    &60.00\std{1.49} &36.28\std{6.00}  &52.80\std{2.72} &41.42\std{4.48} &51.49\std{6.44} &48.40\std{4.23}\\
&\linegrey{TimesNet} &60.67\std{2.91} &39.35\std{6.69} &52.91\std{3.66} &42.00\std{6.52} &54.63\std{6.25} &49.91\std{5.21} \\
&\linegrey{ModernTCN} &59.00\std{1.33} &34.75\std{7.16} &51.94\std{3.89} &39.80\std{5.92} &52.57\std{5.14} &47.61\std{4.69} \\
&\linegrey{TimeMixer} &67.00\std{2.21} &65.44\std{6.53} &63.03\std{1.37} &58.10\std{2.79} &64.11\std{3.05} &63.54\std{3.19} \\
&\linegrey{TSMixer} &63.33\std{3.33} &58.11\std{6.74} &58.06\std{5.41} &48.91\std{7.91} &58.11\std{5.39} &57.30\std{5.76} \\
&\linegrey{PatchTST} &64.67\std{1.63} &55.59\std{3.03} &59.54\std{2.06} &52.52\std{2.50} &58.80\std{2.93} &58.22\std{2.43} \\
&\linegrey{iTransformer} &\secondres{70.67}\std{3.89} &71.59\std{6.26} &\secondres{67.20}\std{5.58} &\secondres{63.79}\std{8.01} &\secondres{63.59}\std{7.94} &\secondres{67.37}\std{6.34} \\
&\linegrey{Leddam} &70.33\std{3.23} &\secondres{72.35}\std{3.41} &67.14\std{5.33} &63.17\std{6.65} &63.31\std{9.18} &67.26\std{5.56} \\
\linepurple{&DeCI &\boldres{73.67}\std{1.64} &\boldres{73.27}\std{2.70} &\boldres{70.34}\std{2.44} &\boldres{67.21}\std{3.01} &\boldres{66.63}\std{4.77} &\boldres{70.22}\std{2.91} \\}

\midrule
\multirow{18}{*}{\rotatebox{90}{\textbf{Neurocon}}}
&\lineorange{SVM} &60.33\std{4.85}&59.16\std{5.30}&59.47\std{5.32}&52.50\std{7.02}&56.93\std{6.43}&57.68\std{5.78}  \\
&\lineorange{Random Forest} &61.00\std{6.00}&58.62\std{9.00}&60.93\std{7.12}&56.85\std{7.47}&58.18\std{9.02}&59.12\std{7.72}  \\
&\lineorange{BrainOOD}  &66.33\std{4.00}&40.69\std{9.89}&54.27\std{5.81}&44.86\std{8.32}&61.56\std{9.36}&53.54\std{8.08}  \\
&\lineorange{BrainNetTF}  &75.11\std{3.65}&74.14\std{8.24}&69.87\std{4.53}&66.64\std{6.46}&72.49\std{6.05}&71.65\std{5.79}  \\
&\linegreen{BrainTGL}  &65.33\std{2.45} &36.50\std{5.92}  &52.67\std{3.27} &42.53\std{4.60} &45.16\std{5.25} &48.44\std{4.30}\\
&\linegreen{MDGL}      &63.83\std{1.00} &33.85\std{4.36}  &50.67\std{1.33} &39.90\std{2.26} &53.82\std{3.99} &48.41\std{2.59}\\
&\linegreen{MVHO}      &68.22\std{2.99} &50.85\std{7.84}  &60.20\std{4.28} &52.60\std{5.29} &59.98\std{5.69} &58.37\std{5.22}\\
&\linegreen{SimMVF}    &52.67\std{9.13}&35.03\std{9.80} &51.07\std{3.26} &38.19\std{9.66} &51.07\std{3.26} &45.61\std{8.42}\\
&\linegreen{PSCRAttn}  &69.83\std{2.55} &51.17\std{4.52}  &60.00\std{3.84} &52.68\std{4.61} &60.84\std{7.76} &58.90\std{4.66}\\
&\linegreen{STAGIN}    &63.83\std{1.00} &33.35\std{3.37}  &50.93\std{1.87} &40.16\std{2.79} &52.98\std{8.41} &48.25\std{4.29}\\
&\linegrey{TimesNet} &64.78\std{1.90}&38.19\std{8.69}&52.00\std{2.67}&42.14\std{4.49}&52.18\std{2.92}&49.86\std{4.13}  \\
&\linegrey{ModernTCN} &63.33\std{1.56}&61.67\std{2.22}&50.00\std{3.14}&38.77\std{1.03}&50.00\std{2.45}&52.75\std{2.08}  \\
&\linegrey{TimeMixer} &72.78\std{6.57}&58.56\std{7.83}&65.13\std{6.72}&59.78\std{5.13}&64.33\std{8.94}&64.12\std{7.04}  \\
&\linegrey{TSMixer} &73.11\std{1.67}&64.53\std{8.36}&64.67\std{2.56}&59.97\std{3.96}&62.31\std{5.66}&64.92\std{4.44}  \\
&\linegrey{PatchTST} &69.33\std{5.61}&54.50\std{8.27}&60.13\std{7.05}&53.59\std{5.78}&59.96\std{6.33}&59.50\std{6.61}  \\
&\linegrey{iTransformer} &82.78\std{2.44}&80.84\std{4.91}&77.07\std{2.44}&75.81\std{2.94}&77.33\std{6.72}&78.77\std{3.89}  \\
&\linegrey{Leddam} &\secondres{83.56}\std{1.88}&\secondres{84.72}\std{5.09}&\secondres{80.27}\std{2.73}&\secondres{80.13}\std{2.74}&\secondres{78.62}\std{3.38}&\secondres{81.46}\std{3.16}  \\
\linepurple{&DeCI&\boldres{87.67}\std{1.75}&\boldres{86.11}\std{3.41}&\boldres{84.13}\std{2.44}&\boldres{82.79}\std{2.49}&\boldres{83.02}\std{2.94}&\boldres{84.74}\std{2.61}  \\}

\midrule
\multirow{18}{*}{\rotatebox{90}{\textbf{PPMI}}}
&\lineorange{SVM} &60.34\std{0.80} &16.33\std{1.27} &24.67\std{0.40} &19.23\std{0.52} &63.93\std{2.76} &36.90\std{1.15} \\
&\lineorange{Random Forest} &58.13\std{1.85} &18.61\std{4.28} &24.55\std{1.54} &20.25\std{2.35} &47.51\std{1.73} &33.81\std{2.35} \\
&\lineorange{BrainOOD}  &65.52\std{1.51} &36.57\std{4.94} &31.96\std{2.53} &29.06\std{3.18} &64.61\std{3.23} &45.54\std{3.08} \\
&\lineorange{BrainNetTF}  &63.31\std{0.65} &28.22\std{4.04} &29.28\std{2.88} &24.91\std{2.14} &54.83\std{3.82} &40.11\std{2.71} \\
&\linegreen{BrainTGL}  &62.97\std{1.00} &24.66\std{3.53} &26.73\std{0.94} &22.11\std{1.46} &51.05\std{3.09} &37.50\std{2.00}\\
&\linegreen{MDGL}      &61.65\std{0.22} &16.40\std{2.03} &25.13\std{0.25} &19.28\std{0.47} &48.76\std{4.69} &34.24\std{1.53}\\
&\linegreen{MVHO}      &62.65\std{0.93} &24.94\std{2.58} &28.75\std{1.27} &25.12\std{1.61} &52.52\std{0.82} &38.80\std{1.44}\\
&\linegreen{SimMVF}    &68.19\std{1.57} &43.05\std{4.16} &39.32\std{2.18} &37.58\std{2.67} &60.70\std{1.52} &49.77\std{2.42}\\
&\linegreen{PSCRAttn}  &64.52\std{1.29} &29.66\std{7.04} &31.79\std{4.01} &27.51\std{3.56} &57.22\std{6.10} &42.14\std{4.40}\\
&\linegreen{STAGIN}    &61.87\std{0.43} &16.87\std{1.82} &25.46\std{0.60} &19.86\std{1.03} &50.14\std{4.60} &34.84\std{1.70}\\
&\linegrey{TimesNet} &82.46\std{2.67} &61.87\std{4.18} &64.63\std{3.89} &61.45\std{4.66} &87.61\std{1.35} &71.60\std{3.35} \\
&\linegrey{ModernTCN} &85.00\std{1.79} &74.99\std{4.39} &70.62\std{5.04} &70.35\std{4.92} &85.46\std{2.24} &77.28\std{3.68} \\
&\linegrey{TimeMixer} &68.09\std{1.55} &40.38\std{5.27} &40.99\std{7.99} &36.91\std{6.39} &71.24\std{2.88} &51.52\std{4.82} \\
&\linegrey{TSMixer} &63.53\std{1.24} &22.49\std{4.58} &29.60\std{3.14} &24.42\std{3.75} &53.44\std{2.17} &38.70\std{2.98} \\
&\linegrey{PatchTST} &\boldres{91.89}\std{0.59} &\boldres{83.52}\std{2.59} &\boldres{85.24}\std{1.26} &\boldres{84.05}\std{1.97} &\boldres{90.23}\std{0.75} &\boldres{86.99}\std{1.43} \\
&\linegrey{iTransformer} &75.63\std{1.75} &60.22\std{4.12} &55.04\std{3.23} &54.46\std{3.47} &72.88\std{2.84} &63.65\std{3.08} \\
&\linegrey{Leddam} &76.96\std{1.89} &60.81\std{6.28} &58.16\std{3.68} &57.36\std{4.10} &78.68\std{2.58} &66.39\std{3.71} \\
\linepurple{&DeCI &\secondres{89.93}\std{0.82} &\secondres{80.67}\std{2.16} &\secondres{79.62}\std{2.09} &\secondres{79.15}\std{1.86} &\secondres{88.34}\std{0.69} &\secondres{83.54}\std{1.52} \\}
\midrule

\end{tabular}}
\end{threeparttable}
\label{tab:0bench_std}
\end{table*}
\begin{table*}[!h]
\small
\centering
\caption{Performance (mean$_{std}$) of all methods on Five-fold cross-validation across TaoWu, ADNI, and ABIDE-120. All results are obtained over five runs. The ``Avg'' metric means the average value of the five evaluation metrics. The best performance is highlighted in \boldres{red}, and the second is highlighted in \secondres{blue}.}
\begin{threeparttable}
\resizebox{0.86\textwidth}{!}{%
\begin{tabular}{c|rcccccc}
\midrule
\multicolumn{1}{c}{} &Metrics & Accuracy & Precision & Recall & F1-Score & AUROC & Avg\\ 
\midrule
\multirow{18}{*}{\rotatebox{90}{\textbf{TaoWu}}}
&\lineorange{SVM} &53.61\std{4.33} &55.41\std{2.42} &52.14\std{1.33} &53.33\std{5.32} &55.00\std{1.22} &53.90\std{2.92} \\
&\lineorange{Random Forest} &58.50\std{7.52} &60.34\std{8.97} &58.50\std{7.52} &55.94\std{8.56} &61.50\std{6.23} &58.96\std{7.76} \\
&\lineorange{BrainOOD}  &78.00\std{4.85} &81.02\std{6.24} &78.00\std{4.85} &76.30\std{6.04} &80.00\std{2.24} &78.66\std{4.84} \\
&\lineorange{BrainNetTF}  &72.50\std{3.16} &77.52\std{5.08} &72.50\std{3.16} &69.94\std{3.95} &69.00\std{7.31} &72.29\std{4.53} \\
&\linegreen{BrainTGL}  &57.50\std{2.74} &45.90\std{5.34} &57.50\std{2.74} &46.33\std{4.66} &56.50\std{9.87} &52.75\std{5.47}\\
&\linegreen{MDGL}      &52.00\std{2.92} &29.74\std{5.85} &52.00\std{2.92} &36.41\std{4.25} &48.00\std{9.32}  &43.63\std{5.05}\\
&\linegreen{MVHO}      &65.50\std{6.00} &65.98\std{9.33} &65.50\std{6.00} &61.23\std{6.55} &66.87\std{7.07}  &65.02\std{6.99}\\
&\linegreen{SimMVF}    &51.50\std{3.00} &39.92\std{9.39}&51.50\std{3.00} &41.94\std{7.99} &51.50\std{3.00}  &47.27\std{6.08}\\
&\linegreen{PSCRAttn}  &71.00\std{4.64} &76.34\std{6.92} &71.00\std{4.64} &67.24\std{5.87} &67.75\std{6.86}  &70.67\std{5.79}\\
&\linegreen{STAGIN}    &57.50\std{3.54} &43.95\std{7.92} &57.50\std{3.54} &46.09\std{5.36} &57.22\std{4.11}  &52.45\std{4.89}\\
&\linegrey{TimesNet} &64.50\std{3.32} &58.30\std{7.83} &64.50\std{3.32} &56.63\std{4.62} &62.50\std{2.24} &61.29\std{4.27} \\
&\linegrey{ModernTCN} &66.50\std{6.04} &63.71\std{8.78} &66.50\std{6.04} &59.56\std{9.55} &62.88\std{9.70} &63.83\std{8.02} \\
&\linegrey{TimeMixer} &66.00\std{8.60} &60.55\std{8.50} &66.00\std{8.60} &59.05\std{6.77} &66.25\std{8.84} &63.57\std{8.26} \\
&\linegrey{TSMixer} &65.00\std{4.18} &62.78\std{7.14} &65.00\std{4.18} &58.76\std{6.21} &62.50\std{6.80} &62.81\std{5.70} \\
&\linegrey{PatchTST} &56.00\std{6.44} &39.36\std{8.12} &56.00\std{6.44} &42.52\std{6.68} &56.13\std{6.15} &50.00\std{6.77} \\
&\linegrey{iTransformer} &\secondres{84.50}\std{7.31} &\secondres{87.48}\std{7.74} &\secondres{84.50}\std{7.31} &\secondres{83.04}\std{8.77} &\secondres{83.75}\std{8.73} &\secondres{84.65}\std{7.97} \\
&\linegrey{Leddam} &75.00\std{5.00} &77.56\std{8.25} &75.00\std{5.00} &72.34\std{6.39} &74.75\std{7.22} &74.93\std{6.37} \\
\linepurple{&DeCI &\boldres{86.50}\std{2.36} &\boldres{88.93}\std{2.88} &\boldres{86.50}\std{2.36} &\boldres{86.17}\std{3.53} &\boldres{88.50}\std{3.67} &\boldres{87.32}\std{2.96} \\}

\midrule
\multirow{18}{*}{\rotatebox{90}{\textbf{ADNI}}}
&\lineorange{SVM}        &75.72\std{0.67} &48.27\std{0.70} &49.19\std{0.56} &48.49\std{0.61} &\boldres{83.70}\std{0.46} &61.07\std{0.60}\\
&\lineorange{Random Forest}         &66.14\std{0.27} &42.30\std{3.79} &35.06\std{0.28} &29.91\std{0.56} &69.45\std{1.78} &48.57\std{1.34}\\
&\lineorange{BrainOOD}   &\secondres{78.78}\std{0.32} &\secondres{53.14}\std{3.47} &\secondres{51.02}\std{1.16} &\secondres{51.09}\std{1.85} &\secondres{79.42}\std{1.11} &\secondres{62.69}\std{1.58}\\
&\lineorange{BrainNetTF} &71.61\std{2.13} &48.30\std{2.13} &42.97\std{2.94} &41.66\std{3.80} &69.97\std{4.79} &54.90\std{3.16}\\
&\linegreen{BrainTGL}   &67.58\std{1.68} &35.39\std{8.87} &38.57\std{3.04} &33.92\std{4.43} &59.18\std{5.16} &46.93\std{4.64}\\
&\linegreen{MDGL}       &64.75\std{0.06} &22.91\std{2.68} &33.36\std{0.06} &26.26\std{0.13} &50.55\std{1.21} &39.57\std{0.83}\\
&\linegreen{MVHO}       &65.33\std{0.14} &37.25\std{3.66} &34.26\std{0.23} &28.29\std{0.51} &51.06\std{0.35} &43.24\std{0.98}\\
&\linegreen{SimMVF}     &69.50\std{0.72} &48.67\std{1.70} &42.52\std{1.21} &42.23\std{1.67} &58.42\std{1.14} &52.27\std{1.29}\\
&\linegreen{PSCRAttn}   &67.75\std{0.95} &46.69\std{2.99} &38.74\std{0.80} &36.20\std{1.37} &57.62\std{1.35} &49.40\std{1.49}\\
&\linegreen{STAGIN}     &64.72\std{0.27} &21.57\std{1.22} &33.33\std{1.14} &26.19\std{1.82} &50.17\std{1.32} &39.20\std{1.15}\\
&\linegrey{TimesNet}   &71.19\std{3.76} &47.42\std{9.20} &46.27\std{8.82} &43.76\std{8.32}&64.20\std{6.47} &54.57\std{8.11}\\
&\linegrey{ModernTCN}  &74.56\std{2.42} &53.69\std{5.18} &49.87\std{3.53} &48.16\std{3.11} &65.15\std{1.81} &58.29\std{3.21}\\
&\linegrey{TimeMixer}  &65.39\std{0.84} &34.82\std{8.63}&35.52\std{2.78} &29.86\std{4.17} &52.57\std{3.98} &43.63\std{4.88}\\
&\linegrey{TSMixer}    &66.00\std{0.34} &37.52\std{3.91} &35.79\std{0.99} &30.49\std{1.29} &52.15\std{0.74} &44.39\std{1.45}\\
&\linegrey{PatchTST}   &66.39\std{1.98} &39.70\std{5.87} &37.65\std{5.53} &32.31\std{6.26} &53.85\std{4.29} &45.98\std{4.78}\\
&\linegrey{iTransformer}&66.25\std{0.88} &34.70\std{4.97} &35.33\std{1.06} &30.02\std{1.92} &57.33\std{5.52} &44.73\std{2.87}\\
&\linegrey{Leddam}     &68.31\std{0.59} &50.17\std{5.55} &41.75\std{1.67} &40.35\std{2.11} &56.79\std{2.90} &51.47\std{2.56}\\
\linepurple{&DeCI &\boldres{79.56}\std{2.62} &\boldres{64.96}\std{5.62} &\boldres{63.00}\std{6.49} &\boldres{62.48}\std{6.74} &78.25\std{4.29} &\boldres{69.65}\std{5.15}}\\

\midrule
\multirow{18}{*}{\rotatebox{90}{\textbf{ABIDE-120}}}
&\lineorange{SVM} &57.28\std{0.98} &47.62\std{1.22} &54.92\std{1.08} &47.43\std{1.82} &51.01\std{3.66} &51.65\std{1.75} \\
&\lineorange{Random Forest} &53.72\std{3.82} &53.64\std{4.74} &53.40\std{3.94} &52.28\std{4.20} &51.97\std{2.99} &53.00\std{3.94} \\
&\lineorange{BrainOOD}&64.72\std{2.43} &63.10\std{2.75} &64.84\std{2.84} &63.07\std{3.62} &59.88\std{4.28} &63.12\std{3.18} \\
&\lineorange{BrainNetTF}&65.13\std{1.58} &66.56\std{2.53} &65.02\std{1.58} &62.79\std{1.58} &60.40\std{3.37} &63.98\std{2.13}\\
&\linegreen{BrainTGL}  &58.08\std{2.25} &46.28\std{7.14} &55.74\std{2.50} &45.58\std{3.90} &58.45\std{7.23} &52.83\std{4.60}\\
&\linegreen{MDGL}      &54.04\std{1.75} &33.17\std{9.14} &51.28\std{2.20} &37.48\std{4.94} &54.56\std{5.35} &46.11\std{4.68}\\
&\linegreen{MVHO}      &61.01\std{1.65} &67.09\std{3.90} &59.18\std{2.09} &53.37\std{3.89} &60.53\std{3.36} &60.24\std{2.98}\\
&\linegreen{SimMVF}    &51.21\std{1.70} &38.39\std{9.12} &49.69\std{1.84} &41.04\std{7.55} &49.69\std{1.84} &46.00\std{4.81}\\
&\linegreen{PSCRAttn}  &57.34\std{1.48} &59.79\std{5.28} &55.88\std{2.06} &47.97\std{3.52} &50.44\std{6.54} &54.28\std{3.78}\\
&\linegreen{STAGIN}    &54.03\std{1.29} &32.29\std{8.08} &51.14\std{1.44} &36.85\std{2.78} &54.75\std{3.67} &45.81\std{3.45}\\
&\linegrey{TimesNet} &58.25\std{3.75} &42.74\std{5.82} &56.07\std{4.16} &45.52\std{7.26} &58.75\std{6.22} &52.27\std{5.44}\\
&\linegrey{ModernTCN} &57.35\std{3.01} &51.98\std{7.71} &56.50\std{2.68} &47.56\std{5.22} &57.77\std{3.28} &54.23\std{4.38}\\
&\linegrey{TimeMixer} &59.58\std{3.07} &55.42\std{6.36} &57.69\std{3.07} &49.78\std{6.20} &58.99\std{4.76} &56.29\std{4.69}\\
&\linegrey{TSMixer} &59.57\std{2.03} &57.48\std{6.10} &58.15\std{2.76} &51.24\std{5.26} &56.14\std{4.91} &56.52\std{4.21}\\
&\linegrey{PatchTST} &61.33\std{3.51} &61.98\std{7.61} &59.91\std{4.42} &54.08\std{8.77} &58.83\std{4.91} &59.23\std{5.84} \\
&\linegrey{iTransformer} &\secondres{66.27}\std{3.07} &\boldres{71.99}\std{2.23} &\secondres{65.40}\std{3.56} &\secondres{63.28}\std{4.27} &\secondres{62.33}\std{5.21} &\secondres{65.85}\std{3.67}\\
&\linegrey{Leddam} &65.22\std{1.82} &\secondres{71.26}\std{2.83} &64.26\std{2.01} &60.26\std{3.62} &59.53\std{5.64} &64.11\std{3.18}\\
\linepurple{&DeCI &\boldres{66.92}\std{1.73} &69.09\std{1.35} &\boldres{66.21}\std{1.60} &\boldres{64.02}\std{1.40} &\boldres{63.72}\std{1.33} &\boldres{65.99}\std{1.48}\\}
\midrule

\end{tabular}}
\end{threeparttable}
\label{tab:1bench_std}
\end{table*}
\begin{table*}[!h]
\small
\centering
\caption{Performance (mean$_{std}$) of all methods on Five-fold cross-validation across ABIDE-180/240/300. All results are obtained over five runs. The ``Avg'' metric means the average value of the five evaluation metrics for convenient comparison. The best performance is highlighted in \boldres{red}, and the second is highlighted in \secondres{blue}.}
\begin{threeparttable}
\resizebox{0.86\textwidth}{!}{%
\begin{tabular}{c|rcccccc}
\midrule
\multicolumn{1}{c}{} &Metrics & Accuracy & Precision & Recall & F1-Score & AUROC & Avg\\ 
\midrule
\multirow{18}{*}{\rotatebox{90}{\textbf{ABIDE-180}}}
&\lineorange{SVM} &59.24\std{1.15} &59.78\std{1.43} &58.70\std{1.11} &58.53\std{1.05} &55.63\std{0.40} &58.38\std{1.03} \\
&\lineorange{Random Forest} &60.75\std{3.11} &60.73\std{3.83} &59.76\std{3.17} &\secondres{59.11}\std{3.21} &58.79\std{2.46} &59.83\std{3.16} \\
&\lineorange{BrainOOD}&\secondres{62.54}\std{1.94} &60.73\std{1.63} &\secondres{60.24}\std{2.08} &55.87\std{3.42} &59.20\std{4.52} &59.72\std{2.72}\\
&\lineorange{BrainNetTF}&60.76\std{2.37} &\secondres{65.15}\std{3.46} &58.36\std{2.99} &52.10\std{5.53} &59.31\std{3.47} &59.14\std{3.56}\\
&\linegreen{BrainTGL}  &57.28\std{2.41} &49.82\std{8.35} &54.27\std{2.89} &43.79\std{5.63} &\secondres{60.23}\std{3.35} &53.08\std{4.93}\\
&\linegreen{MDGL}      &53.81\std{0.00} &26.91\std{0.00}  &50.00\std{0.00} &34.98\std{0.00} &50.63\std{6.45} &43.27\std{1.29}\\
&\linegreen{MVHO}      &57.20\std{1.52} &55.30\std{7.83}  &54.33\std{2.01} &45.04\std{3.96} &51.46\std{2.88} &52.67\std{3.64}\\
&\linegreen{SimMVF}    &51.78\std{2.99} &43.50\std{8.98}  &49.79\std{1.75} &41.63\std{6.09} &49.79\std{1.75} &47.30\std{4.31}\\
&\linegreen{PSCRAttn}  &57.45\std{0.98} &56.01\std{8.90}  &54.61\std{1.39} &45.89\std{2.44} &52.93\std{2.02} &53.38\std{3.15}\\
&\linegreen{STAGIN}    &54.31\std{0.81} &30.39\std{4.44}  &50.56\std{0.88} &36.15\std{1.65} &51.11\std{1.12} &44.50\std{1.78}\\
&\linegrey{TimesNet} &54.41\std{1.36} &32.51\std{7.53} &50.94\std{1.20} &36.78\std{2.58} &52.60\std{3.59} &45.45\std{3.25} \\
&\linegrey{ModernTCN} &53.90\std{0.16} &41.86\std{6.24} &50.47\std{0.73} &35.87\std{1.35} &50.59\std{0.87} &46.54\std{1.87} \\
&\linegrey{TimeMixer} &57.38\std{1.34} &52.77\std{5.35} &54.80\std{1.71} &45.35\std{4.00} &57.10\std{1.27} &53.48\std{2.73}\\
&\linegrey{TSMixer} &55.68\std{1.10} &46.03\std{4.86} &52.61\std{0.90} &40.94\std{1.39} &52.78\std{3.36} &49.61\std{2.32}\\
&\linegrey{PatchTST} &58.65\std{2.14} &55.97\std{8.05} &56.05\std{2.47} &48.74\std{4.81} &56.24\std{2.55} &55.13\std{4.00} \\
&\linegrey{iTransformer} &61.78\std{0.90} &64.94\std{4.08} &59.77\std{0.68} &55.74\std{1.63} &59.22\std{2.85} &\secondres{60.29}\std{2.03} \\
&\linegrey{Leddam} &61.96\std{1.75} &64.24\std{2.79} &59.95\std{2.28} &55.80\std{4.48} &56.49\std{3.46} &59.69\std{2.95}\\
\linepurple{&DeCI &\boldres{63.99}\std{1.82} &\boldres{66.55}\std{2.25} &\boldres{62.67}\std{2.19} &\boldres{60.99}\std{2.56} &\boldres{62.04}\std{2.58} &\boldres{63.25}\std{2.28} \\}

\midrule
\multirow{18}{*}{\rotatebox{90}{\textbf{ABIDE-240}}}
&\lineorange{SVM}&54.21\std{1.29} &55.96\std{1.64} &49.17\std{1.29} &53.66\std{1.00} &48.22\std{2.05} &52.24\std{1.45} \\
&\lineorange{Random Forest} &53.83\std{2.87} &54.41\std{3.13} &54.02\std{2.88} &53.28\std{2.70} &56.81\std{3.56} &54.47\std{3.03} \\
&\lineorange{BrainOOD} &\secondres{67.17}\std{3.23} &72.16\std{2.46} &\secondres{66.82}\std{3.43} &\secondres{64.49}\std{4.20} &\secondres{67.06}\std{1.95} &\secondres{67.54}\std{3.05} \\
&\lineorange{BrainNetTF} &63.50\std{3.27} &70.82\std{2.41} &63.20\std{3.40} &59.22\std{5.31} &62.00\std{6.13} &63.75\std{4.10} \\
&\linegreen{BrainTGL}  &55.50\std{1.55} &49.90\std{6.39}  &54.74\std{1.49} &45.44\std{3.61} &51.79\std{3.77} &51.47\std{3.36}\\
&\linegreen{MDGL}      &51.50\std{1.33} &27.31\std{3.78}  &50.67\std{1.33} &34.94\std{2.49} &47.72\std{1.15} &42.43\std{2.02}\\
&\linegreen{MVHO}      &60.17\std{2.20} &63.99\std{7.18}  &59.62\std{2.10} &54.05\std{2.69} &56.84\std{2.34} &58.93\std{3.30}\\
&\linegreen{SimMVF}    &51.17\std{2.67} &45.36\std{9.05}  &51.03\std{2.84} &42.71\std{9.00} &51.03\std{2.84} &48.26\std{5.68}\\
&\linegreen{PSCRAttn}  &61.33\std{1.35} &63.94\std{5.74}  &60.86\std{1.77} &55.06\std{3.57} &60.93\std{4.86} &60.42\std{3.46}\\
&\linegreen{STAGIN}    &52.33\std{2.00} &30.62\std{6.58}  &51.87\std{2.35} &37.16\std{4.61} &46.49\std{3.11} &43.69\std{3.73}\\
&\linegrey{TimesNet}&57.67\std{3.99} &60.03\std{9.19} &57.38\std{3.66} &48.72\std{6.77} &58.30\std{3.35} &56.42\std{5.39} \\
&\linegrey{ModernTCN}&56.33\std{3.71} &59.31\std{8.73} &56.50\std{3.89} &46.89\std{7.94} &56.28\std{4.75} &55.06\std{5.80} \\
&\linegrey{TimeMixer}&59.00\std{1.62} &60.86\std{6.62} &59.08\std{1.83} &51.83\std{2.15} &59.56\std{2.39} &58.07\std{2.92} \\
&\linegrey{TSMixer} &56.67\std{2.53} &50.57\std{7.70} &55.83\std{2.53} &45.95\std{4.44} &56.41\std{1.80} &53.09\std{3.80} \\
&\linegrey{PatchTST}&58.83\std{2.21} &53.31\std{5.40} &58.00\std{2.21} &50.37\std{4.70} &57.86\std{2.48} &55.67\std{3.40} \\
&\linegrey{iTransformer}&66.83\std{3.18} &72.21\std{2.59} &66.79\std{3.03} &64.39\std{4.67} &64.67\std{3.31} &66.98\std{3.36} \\
&\linegrey{Leddam}&66.50\std{2.44} &\secondres{72.70}\std{0.57} &66.30\std{2.79} &63.53\std{4.17} &64.08\std{1.10} &66.62\std{2.21} \\
\linepurple{&DeCI&\boldres{71.33}\std{2.15} &\boldres{75.31}\std{2.88} &\boldres{71.14}\std{2.00} &\boldres{69.79}\std{2.68} &\boldres{72.15}\std{2.17} &\boldres{71.94}\std{2.38} \\}

\midrule
\multirow{18}{*}{\rotatebox{90}{\textbf{ABIDE-300}}}
&\lineorange{SVM}&64.20\std{1.70} &65.96\std{1.12} &62.01\std{1.83} &61.20\std{2.07} &\secondres{64.93}\std{1.14} &63.66\std{1.57} \\
&\lineorange{Random Forest} &54.05\std{3.30} &52.83\std{3.95} &52.97\std{3.65} &52.27\std{4.07} &53.63\std{3.74} &53.15\std{3.74} \\
&\lineorange{BrainOOD}&61.33\std{3.55} &56.07\std{9.68} &57.07\std{4.39} &49.15\std{7.58} &57.90\std{4.48} &56.30\std{5.94} \\
&\lineorange{BrainNetTF} &65.22\std{2.33} &70.61\std{1.41} &61.93\std{2.57} &58.57\std{3.30} &61.34\std{6.72} &63.53\std{3.27} \\
&\linegreen{BrainTGL}  &60.59\std{3.70} &50.36\std{9.45}  &57.20\std{4.79} &49.89\std{8.31} &59.41\std{6.92} &55.49\std{6.63}\\
&\linegreen{MDGL}      &56.94\std{1.40} &31.79\std{4.83}  &51.46\std{1.80} &38.30\std{3.07} &44.56\std{2.92} &44.61\std{2.80}\\
&\linegreen{MVHO}      &59.58\std{2.05} &46.03\std{8.12}  &55.63\std{3.07} &46.75\std{5.85} &57.72\std{2.92} &53.14\std{4.40}\\
&\linegreen{SimMVF}    &54.16\std{3.27} &48.66\std{8.95}  &51.07\std{1.94} &42.77\std{5.13} &51.07\std{1.94} &49.55\std{4.25}\\
&\linegreen{PSCRAttn}  &60.14\std{2.31} &50.24\std{9.80}  &55.90\std{2.74} &47.32\std{5.44} &56.35\std{3.91} &53.99\std{4.84}\\
&\linegreen{STAGIN}    &56.68\std{1.78} &29.98\std{4.16}  &51.17\std{2.33} &37.49\std{3.36} &45.26\std{3.07} &44.12\std{2.94}\\
&\linegrey{TimesNet}&55.79\std{0.00} &27.90\std{0.00} &50.00\std{0.00} &35.81\std{0.00} &50.00\std{0.00} &43.90\std{0.00} \\
&\linegrey{ModernTCN}&55.83\std{1.38} &34.05\std{6.36} &51.51\std{1.48} &38.40\std{2.70} &51.12\std{1.88} &46.18\std{2.76} \\
&\linegrey{TimeMixer}&60.14\std{3.33} &52.64\std{5.68} &56.22\std{3.73} &47.76\std{7.18} &59.24\std{5.71} &55.20\std{5.13} \\
&\linegrey{TSMixer} &58.12\std{2.66} &39.52\std{8.20} &52.80\std{3.22} &40.94\std{5.85} &52.31\std{3.57} &48.74\std{4.70} \\
&\linegrey{PatchTST}&58.99\std{2.83} &58.19\std{6.09} &54.94\std{2.84} &46.03\std{6.49} &54.23\std{2.03} &54.48\std{4.06} \\
&\linegrey{iTransformer}&64.32\std{1.07} &69.44\std{3.71} &61.07\std{0.86} &57.38\std{1.58} &56.15\std{2.66} &61.67\std{1.98} \\
&\linegrey{Leddam}&\secondres{67.54}\std{1.31} &\secondres{71.94}\std{2.72} &\secondres{65.16}\std{1.69} &\secondres{63.67}\std{2.34} &64.50\std{3.92} &\secondres{66.56}\std{2.40} \\
\linepurple{&DeCI&\boldres{69.47}\std{1.60} &\boldres{72.86}\std{1.77} &\boldres{67.19}\std{1.98} &\boldres{65.30}\std{2.01} &\boldres{66.68}\std{2.36} &\boldres{68.30}\std{1.94} \\}
\midrule

\end{tabular}}
\end{threeparttable}
\label{tab:2bench_std}
\end{table*}
\clearpage
\begin{table*}[!p]
\small
\centering
\caption{Beyond AAL atlas (116 ROIs)~\cite{tzourio2002aal116}, we additionally evaluated Schaefer atlas (100 ROIs)~\cite{schaefer2018localglobal}, Ward atlas (100 ROIs)~\cite{Thirion2014ward100}, and Harvard atlas (48 ROIs)~\cite{HarvardOxford2012havard48} to assess parcellation robustness on ABIDE-240 and PPMI. We choose the FC-based \textbf{BrainNetTF}~\cite{kan2022BrainNetTF}, dFC-based \textbf{BrainTGL}~\cite{liu2023braintgldfc2}, overall second-best \textbf{Leddam}~\cite{yu2024leddam}, and our \textbf{DeCI}.}
\begin{threeparttable}
\resizebox{1.0\textwidth}{!}{%
\begin{tabular}{c|rrcccccc}
\midrule
\multicolumn{1}{c}{Dataset}& Models &Atlas & Accuracy & Precision & Recall & F1-Score & AUROC & Avg\\ 
\midrule
\multirow{24}{*}{\rotatebox{90}{\textbf{ABIDE-240}}}
&\multirow{4}{*}{\textbf{BrainNetTF}}
&AAL-116      &63.50\std{3.27} &70.82\std{2.41} &63.20\std{3.40} &59.22\std{5.31} &62.00\std{6.13} &63.75\std{4.10}\\
&&Schaefer-100&63.83\std{3.01} &69.32\std{3.71} &63.48\std{3.17} &60.75\std{3.64} &59.65\std{5.70} &63.41\std{3.85}\\
&&Ward-100    &61.67\std{0.91} &65.22\std{5.11} &61.77\std{1.01} &57.85\std{2.86} &59.75\std{5.53} &61.25\std{3.08}\\
&&Harvard-48  &65.00\std{3.54} &70.60\std{6.95} &64.66\std{3.83} &61.06\std{5.25} &65.90\std{2.03} &65.44\std{4.32}\\\cline{2-9}

&\multirow{4}{*}{\textbf{BrainTGL}}
&AAL-116      &55.50\std{1.55} &49.90\std{6.39} &54.74\std{1.49} &45.44\std{3.61} &51.79\std{3.77} &51.47\std{3.36}\\
&&Schaefer-100&55.33\std{2.21} &52.29\std{7.42} &55.33\std{1.85} &44.82\std{3.84} &55.10\std{1.27} &52.57\std{3.32}\\
&&Ward-100    &55.33\std{1.72} &52.17\std{5.83} &54.76\std{1.98} &44.46\std{4.45} &54.65\std{4.78} &52.27\std{3.75}\\
&&Harvard-48  &54.17\std{1.39} &46.48\std{8.34} &53.35\std{1.42} &41.69\std{3.19} &51.98\std{5.14} &49.53\std{4.30}\\\cline{2-9}

&\multirow{4}{*}{\textbf{Leddam}}
&AAL-116      &66.50\std{2.44} &72.70\std{0.57} &66.30\std{2.79} &63.53\std{4.17} &64.08\std{1.10} &66.62\std{2.21}\\
&&Schaefer-100&67.50\std{1.90} &72.40\std{2.81} &67.23\std{1.82} &65.46\std{2.53} &65.75\std{3.70} &67.67\std{2.55}\\
&&Ward-100    &67.67\std{2.71} &70.02\std{2.20} &67.38\std{2.60} &66.45\std{3.00} &64.61\std{2.30} &67.23\std{2.56}\\
&&Harvard-48  &67.17\std{1.45} &72.74\std{2.38} &66.97\std{1.61} &64.77\std{1.91} &64.92\std{5.15} &67.31\std{2.50}\\\cline{2-9}

&\multirow{4}{*}{\textbf{DeCI}}
&AAL-116      &71.33\std{2.15} &75.31\std{2.88} &71.14\std{2.00} &69.79\std{2.68} &72.15\std{2.17} &71.94\std{2.38}\\
&&Schaefer-100&70.17\std{4.13} &72.83\std{3.67} &69.99\std{4.15} &68.71\std{4.81} &67.69\std{6.04} &69.88\std{4.56}\\
&&Ward-100    &71.00\std{4.36} &72.56\std{3.82} &70.93\std{4.59} &70.26\std{5.00} &69.58\std{5.48} &70.87\std{4.65}\\
&&Harvard-48  &71.50\std{2.95} &75.98\std{2.79} &71.40\std{2.86} &69.69\std{3.57} &72.43\std{3.05} &72.20\std{3.04}\\

\midrule
\multirow{24}{*}{\rotatebox{90}{\textbf{PPMI}}}
&\multirow{4}{*}{\textbf{BrainNetTF}}
&AAL-116      &63.31\std{0.65} &28.22\std{4.04} &29.28\std{2.88} &24.91\std{2.14} &54.83\std{3.82} &40.11\std{2.71} \\
&&Schaefer-100&63.74\std{0.70} &27.66\std{6.21} &28.50\std{1.40} &24.66\std{2.26} &49.01\std{3.65} &38.71\std{2.84}\\
&&Ward-100    &70.40\std{2.17} &36.02\std{4.02} &37.78\std{4.95} &34.58\std{4.53} &65.63\std{4.00} &48.88\std{3.93}\\
&&Harvard-48  &63.30\std{0.63} &23.32\std{6.33} &28.73\std{1.98} &23.68\std{1.45} &54.86\std{4.00} &38.78\std{2.88}\\\cline{2-9}

&\multirow{4}{*}{\textbf{BrainTGL}}
&AAL-116      &62.97\std{1.00} &24.66\std{3.53} &26.73\std{0.94} &22.11\std{1.46} &51.05\std{3.09} &37.50\std{2.00}\\
&&Schaefer-100&62.65\std{0.36} &23.57\std{2.44} &26.40\std{0.57} &21.47\std{0.82} &53.58\std{1.55} &37.53\std{1.15}\\
&&Ward-100    &62.75\std{0.81} &21.85\std{4.68} &27.95\std{0.98} &22.85\std{0.88} &52.52\std{1.58} &37.58\std{1.79}\\
&&Harvard-48  &62.32\std{0.44} &22.05\std{3.79} &26.34\std{0.91} &21.35\std{1.46} &48.65\std{2.94} &36.14\std{1.91}\\\cline{2-9}

&\multirow{4}{*}{\textbf{Leddam}}
&AAL-116      &76.96\std{1.89} &60.81\std{6.28} &58.16\std{3.68} &57.36\std{4.10} &78.68\std{2.58} &66.39\std{3.71} \\
&&Schaefer-100&80.59\std{1.14} &67.41\std{3.81} &64.74\std{3.28} &64.29\std{3.46} &81.66\std{2.14} &71.74\std{2.77}\\
&&Ward-100    &77.35\std{3.84} &65.43\std{8.03} &57.58\std{8.03} &58.05\std{8.30} &79.32\std{3.13} &67.55\std{6.27}\\
&&Harvard-48  &78.58\std{4.24} &64.46\std{8.92} &63.30\std{8.46} &61.59\std{8.55} &79.71\std{6.34} &69.53\std{8.50}\\\cline{2-9}

&\multirow{4}{*}{\textbf{DeCI}}
&AAL-116      &89.93\std{0.82} &80.67\std{2.16} &79.62\std{2.09} &79.15\std{1.86} &88.34\std{0.69} &83.54\std{1.52} \\
&&Schaefer-100&88.51\std{1.11} &79.60\std{3.06} &78.43\std{3.42} &77.92\std{3.58} &86.98\std{2.48} &82.29\std{2.73}\\
&&Ward-100    &91.46\std{0.93} &84.22\std{2.06} &83.84\std{1.91} &82.79\std{1.89} &90.35\std{0.80} &86.53\std{1.52}\\
&&Harvard-48  &89.18\std{1.19} &78.44\std{4.20} &78.06\std{3.09} &77.21\std{3.44} &87.71\std{2.10} &82.12\std{2.80}\\

\midrule

\end{tabular}}
\end{threeparttable}
\label{tab:atlas}
\end{table*}
\begin{table*}[t]
\centering
\caption{Statistical significance analysis of DeCI vs.\ Leddam on four fMRI datasets (Accuracy and F1-Score, $n{=}5$ folds). $\Delta$ denotes mean paired improvement with 95\% CI. $d_z$ is Cohen's $d$ for paired designs ($d_z=\bar d/s_d=t/\sqrt{n}$). As a rule of thumb for $n{=}5$, $d_z>1.2$ indicates strong and stable improvement (typically $p<0.05$). $^{***}p<0.001$, $^{*}p<0.05$. Paired two-sided $t$-tests.}
\label{tab:significance}
\resizebox{1\textwidth}{!}{%
\begin{tabular}{lcccccccc}
\toprule
\multirow{2}{*}{Dataset} & \multicolumn{4}{c}{Accuracy} & \multicolumn{4}{c}{F1-Score} \\
\cmidrule(lr){2-5} \cmidrule(lr){6-9}
 & Leddam & DeCI & $\Delta$/[95\% CI]/$p$ & $d_z$ & Leddam & DeCI & $\Delta$/[95\% CI]/$p$ & $d_z$ \\
\midrule
ADNI     & 68.31$\pm$0.59 & 79.56$\pm$2.62 & +11.25 [8.24, 14.28]$^{***}$  & 4.63 
         & 40.35$\pm$2.11 & 62.48$\pm$6.74 & +22.13 [13.92, 30.33]$^{***}$ & 3.35 \\
Mātai    & 70.33$\pm$3.23 & 73.67$\pm$4.64 & +3.34 [0.35, 6.31]$^{*}$      & 1.39 
         & 63.17$\pm$6.65 & 67.21$\pm$6.01 & +4.04 [0.32, 7.76]$^{*}$      & 1.35 \\
Neurocon & 83.56$\pm$1.88 & 87.67$\pm$1.75 & +4.11 [0.75, 7.47]$^{*}$      & 1.81 
         & 80.13$\pm$2.74 & 82.79$\pm$4.49 & +2.66 [0.18, 5.14]$^{*}$      & 1.33 \\
TaoWu    & 75.00$\pm$5.00 & 86.50$\pm$2.36 & +11.50 [7.96, 15.04]$^{***}$  & 4.03 
         & 72.34$\pm$6.39 & 86.17$\pm$3.53 & +13.83 [9.71, 17.93]$^{***}$  & 4.17 \\ 
\bottomrule
\end{tabular}
}
\end{table*}
\clearpage

\subsection{Ablation Study}

\begin{figure}[t]
  \centering
  \includegraphics[width=0.48\textwidth]{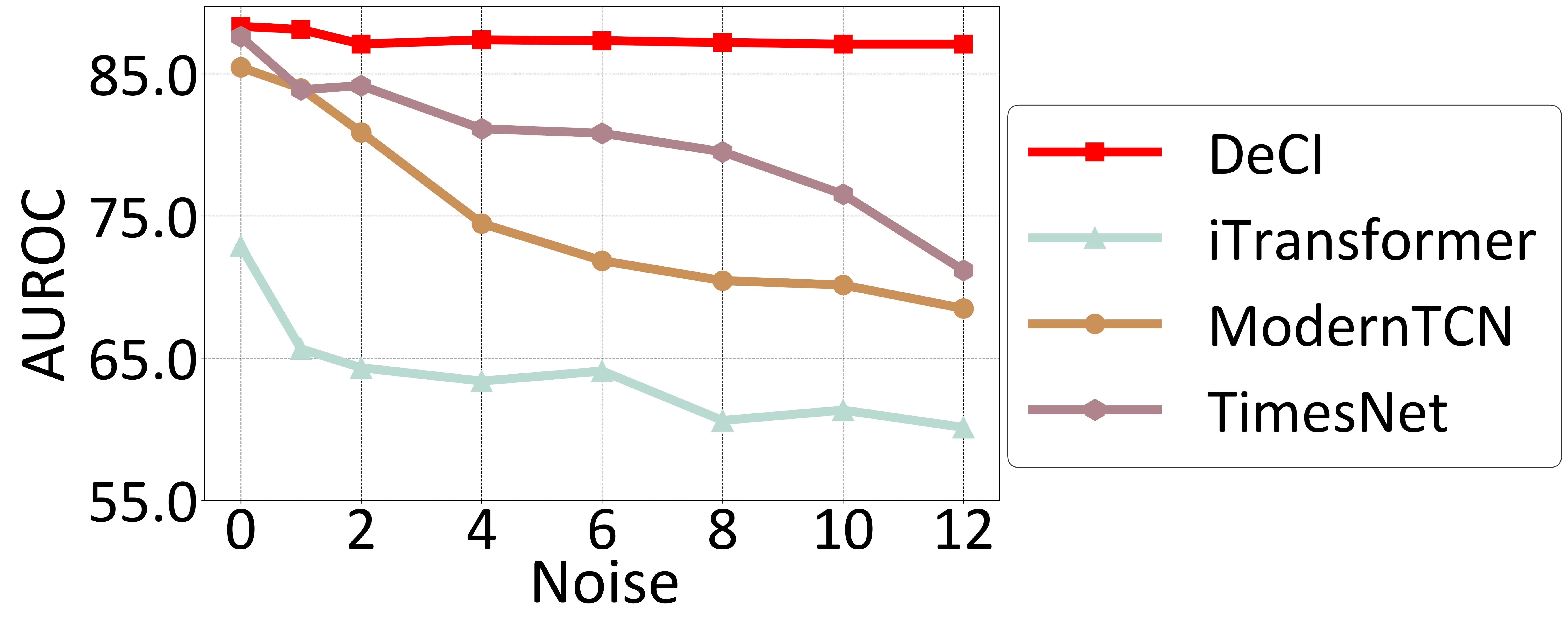}
\caption{Test of resilience to channel noise on PPMI. DeCI is Channel-Independent; others are Channel-Dependent. Gaussian noise with zero mean and varying intensity ($\beta \in {0, 1, \dots, 12}$) is added to the last half of the training channels.}
  \label{fig:noise}
\end{figure}

\subsubsection{Analysis of Channel-Independent Strategy} 

Compared to the Channel-Dependent (CD) method, the Channel-Independent (CI) strategy helps models preserve the unique patterns of each channel (ROI), improving robustness to channel noise. Fig.~\ref{fig:noise} highlights that CI models like our DeCI maintain robust performance under noise perturbations, while CD models (all three other baselines) rapidly degrade and show high sensitivity to noise from other channels. 

\begin{figure}[t]
  \centering
  \includegraphics[width=0.48\textwidth]{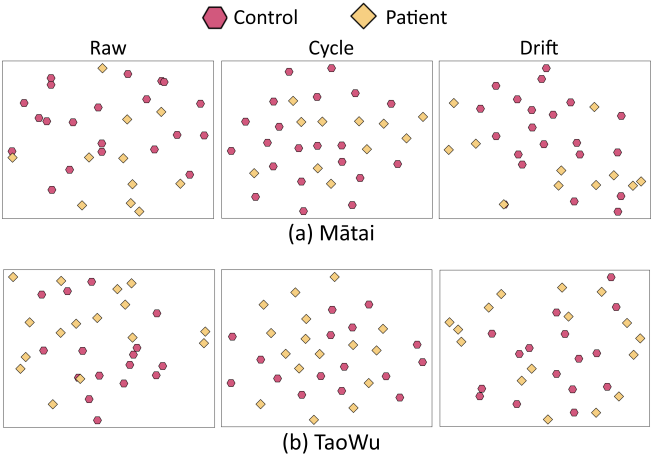}
  \caption{Study of Cycle and Drift Decomposition on Mātai and TaoWu dataset. We plot the t-SNE visualization of the raw series, the $Cycle$ and $Drift$ patterns. We also evaluate the discriminative capabilities of Logistic Regression (LR) and Support Vector
Machines (SVM) on different series, using AUROC as an indicator, with the higher value indicating the better performance.}
  \label{fig:emb_std}
\end{figure}

\begin{table}[t]
\small
\centering
\caption{Evaluation of Support Vector Machine (SVM) and Logistic Regression (LR) on raw BOLD signals, Cycle patterns, and Drift patterns. We use AUROC to quantify the performance; the higher, the better.}
\vspace{-2mm}
\resizebox{0.48\textwidth}{!}{
\begin{threeparttable}
\begin{tabular}{cccccccc}
\midrule
 &\multicolumn{3}{c}{Mātai}&\multicolumn{3}{c}{TaoWu}\\ 
\cmidrule(l{10pt}r{10pt}){2-4}\cmidrule(l{10pt}r{10pt}){5-7}

 & Raw & Cycle & Drift & Raw & Cycle & Drift\\ 
\midrule
SVM &0.552    &0.596     &0.634  &0.619    &0.809     &0.744\\
LR	&0.626    &0.761     &0.714  &0.619    &0.809     &0.744 \\
\midrule
\end{tabular}
\end{threeparttable}
}
\label{tab:std}
\vspace{-2mm}
\end{table}
\begin{figure}[!t]
\includegraphics[width=0.49\textwidth]{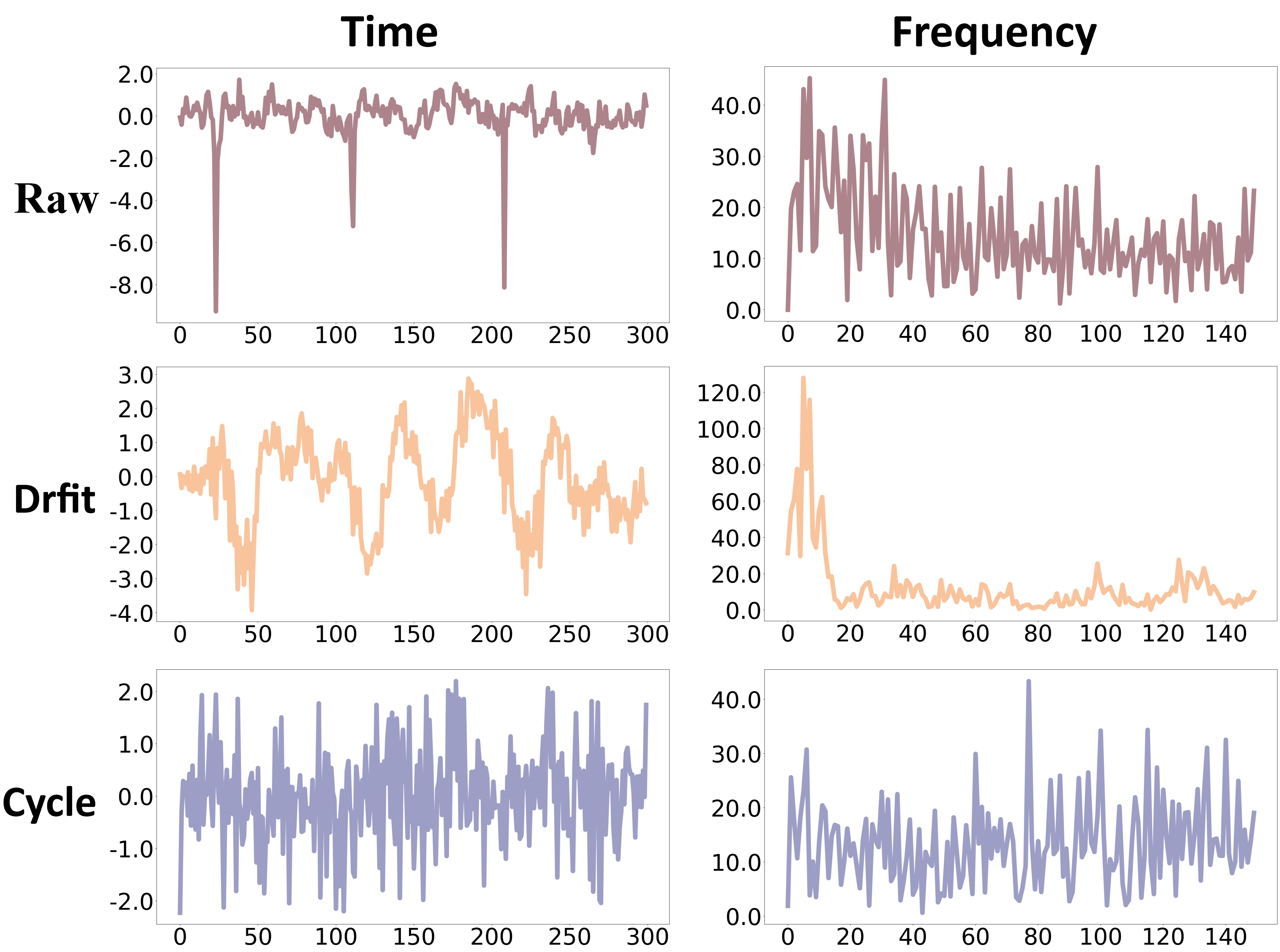}
\caption{Visualization of decomposed drift and cycle patterns for the 24th ROI from ABIDE. The Drift component captures the relatively lower-frequency part of the band-limited signals, whereas the Cycle component represents the relatively higher-frequency part.}
\label{fig:visual_cd}
\end{figure}

\subsubsection{Analysis of Cycle and Drift Decomposition}
Fig.~\ref{fig:emb_std} shows t-SNE visualizations of the raw time series and its Cycle and Drift components. In the raw embedding space, class overlap is substantial, impeding classification. However, embeddings of different classes from the Cycle and Drift patterns are clearly separated, highlighting that Cycle and Drift Decomposition generates more discriminative features, as confirmed by the performance of LR and SVM in Table~\ref{tab:std}.
Qualitative plots of learned Drift (relatively low-frequency) and Cycle (relatively high-frequency) in Fig.~\ref{fig:visual_cd} show slow baseline drift and oscillatory cycle being adaptively separated by the proposed Cycle and Drift Decomposition.
Table~\ref{tab:std_ablation} further illustrates the effectiveness of our progressive decomposition scheme in DeCI. 
The variant without any pattern extraction performs the worst across nearly all datasets. Extracting either the Cycle or the Drift component individually results in significant improvements, while jointly capturing both components (DeCI) yields the best performance overall. 

Additionally, we conducted a sensitivity analysis on the kernel size $K$ of the 1D convolution used to extract the Drift component. As shown in Table~\ref{tab:kernel}, setting $K=64$ (equal to the model dimension $D=64$, ensuring a full receptive field) yields the best performance on both datasets. Smaller or mid-range kernels ($K=56$–$32$) are generally under-smooth and impair performance. We therefore recommend the default setting $K = D$.

\subsubsection{Analysis of Channel-Independent Strategy}
We conducted an ablation of the Channel-Independent (CI) strategy by introducing inter-ROI interaction modeling either during representation learning (\textit{inter-CD}) or only at the final aggregation of ROI representations (\textit{final-CD}). As shown in Table~\ref{tab:ci_ablation}, adding Channel-Dependence at either stage yields a significant performance degradation. This confirms that the CI strategy enhances robustness by preserving unique ROI patterns, while also mitigating overfitting to spurious inter-ROI dependencies.
\begin{table}[t]
\centering
\def\arraystretch{1.0}
\vspace{-2mm}
\caption{Ablation of the proposed Cycle and Drift Decomposition. \textit{(i)} no decomposition is not performed; \textit{(ii)} only $Cycle$ information is extracted and used; \textit{(iii)} only $Drift$ information is extracted and used; \textit{(iv)} Both.}
\vspace{-2mm}
\label{tab:std_ablation}
\resizebox{0.45\textwidth}{!}{
\begin{tabular}{cccccccc}

    \toprule
    \scalebox{2.5}{\textbf{ }} & \scalebox{2.5}{\textbf{Cycle}} & \scalebox{2.5}{\textbf{Drift}} & \multicolumn{1}{c}{\scalebox{2.5}{\textbf{Accuracy
    }}} & \multicolumn{1}{c}{\scalebox{2.5}{\textbf{Precision}}} & \multicolumn{1}{c}{\scalebox{2.5}{\textbf{Recall}}} & \multicolumn{1}{c}{\scalebox{2.5}{\textbf{F1-Score}}} & \multicolumn{1}{c}{\scalebox{2.5}{\textbf{AUROC}}} \\

    \midrule
    \multirow{4}{*}{\scalebox{2.5}{\rotatebox{90}{\textbf{ Mātai}}}}

&\scalebox{5}{-} & \scalebox{5}{-}
  & \scalebox{2.5}{64.00}
  & \scalebox{2.5}{54.95}
  & \scalebox{2.5}{58.29}
  & \scalebox{2.5}{51.41}
  & \scalebox{2.5}{56.80}  \\

& \scalebox{4}{\checkmark} & \scalebox{5}{-}
  & \scalebox{2.5}{69.33}
  & \scalebox{2.5}{70.63}
  & \scalebox{2.5}{65.14}
  & \scalebox{2.5}{62.85}
  & \scalebox{2.5}{62.06}  \\

& \scalebox{5}{-} & \scalebox{4}{\checkmark}
  & \scalebox{2.5}{71.00}
  & \scalebox{2.5}{70.87}
  & \scalebox{2.5}{67.71}
  & \scalebox{2.5}{64.34}
  & \scalebox{2.5}{64.46}  \\

& \scalebox{4}{\checkmark} & \scalebox{4}{\checkmark}
  & \boldres{\scalebox{2.5}{73.67}}
  & \boldres{\scalebox{2.5}{73.27}}
  & \boldres{\scalebox{2.5}{70.34}}
  & \boldres{\scalebox{2.5}{67.21}}
  & \boldres{\scalebox{2.5}{66.63}}  \\

\midrule
    \multirow{4}{*}{\scalebox{2.5}{\rotatebox{90}{\textbf{ TaoWu}}}}

&\scalebox{5}{-} & \scalebox{5}{-}
  & \scalebox{2.5}{68.00}
  & \scalebox{2.5}{68.18}
  & \scalebox{2.5}{68.00}
  & \scalebox{2.5}{63.02}
  & \scalebox{2.5}{67.75}  \\

& \scalebox{4}{\checkmark} & \scalebox{5}{-}
  & \scalebox{2.5}{77.50}
  & \scalebox{2.5}{79.74}
  & \scalebox{2.5}{77.50}
  & \scalebox{2.5}{76.10}
  & \scalebox{2.5}{74.50}  \\

& \scalebox{5}{-} & \scalebox{4}{\checkmark}
  & \scalebox{2.5}{82.00}
  & \scalebox{2.5}{87.30}
  & \scalebox{2.5}{82.00}
  & \scalebox{2.5}{80.54}
  & \scalebox{2.5}{84.75}  \\
& \scalebox{4}{\checkmark} & \scalebox{4}{\checkmark}
  & \boldres{\scalebox{2.5}{86.50}}
  & \boldres{\scalebox{2.5}{88.93}}
  & \boldres{\scalebox{2.5}{86.50}}
  & \boldres{\scalebox{2.5}{86.17}}
  & \boldres{\scalebox{2.5}{88.50}}  \\

    \midrule
    \multirow{4}{*}{\scalebox{2.5}{\rotatebox{90}{\textbf{ Neurocon}}}}

&\scalebox{5}{-} & \scalebox{5}{-}
  & \scalebox{2.5}{70.61}
  & \scalebox{2.5}{56.90}
  & \scalebox{2.5}{62.47}
  & \scalebox{2.5}{55.93}
  & \scalebox{2.5}{60.22}  \\

& \scalebox{4}{\checkmark} & \scalebox{5}{-}
  & \scalebox{2.5}{80.22}
  & \scalebox{2.5}{79.83}
  & \scalebox{2.5}{76.07}
  & \scalebox{2.5}{74.30}
  & \scalebox{2.5}{72.80}  \\

& \scalebox{5}{-} & \scalebox{4}{\checkmark}
  & \scalebox{2.5}{81.06}
  & \scalebox{2.5}{79.02}
  & \scalebox{2.5}{75.87}
  & \scalebox{2.5}{74.28}
  & \scalebox{2.5}{76.76}  \\

& \scalebox{4}{\checkmark} & \scalebox{4}{\checkmark}
  & \boldres{\scalebox{2.5}{87.67}}
  & \boldres{\scalebox{2.5}{86.11}}
  & \boldres{\scalebox{2.5}{84.13}}
  & \boldres{\scalebox{2.5}{82.79}}
  & \boldres{\scalebox{2.5}{83.02}}  \\

    \midrule
    \multirow{4}{*}{\scalebox{2.5}{\rotatebox{90}{\textbf{ PPMI}}}}
&\scalebox{5}{-} & \scalebox{5}{-}
  & \scalebox{2.5}{79.40}
  & \scalebox{2.5}{71.36}
  & \scalebox{2.5}{70.75}
  & \scalebox{2.5}{69.68}
  & \scalebox{2.5}{77.76}  \\

& \scalebox{4}{\checkmark} & \scalebox{5}{-}
  & \scalebox{2.5}{87.01}
  & \scalebox{2.5}{77.94}
  & \scalebox{2.5}{77.09}
  & \scalebox{2.5}{76.72}
  & \scalebox{2.5}{86.93}  \\

& \scalebox{5}{-} & \scalebox{4}{\checkmark}
  & \scalebox{2.5}{83.81}
  & \scalebox{2.5}{71.34}
  & \scalebox{2.5}{69.28}
  & \scalebox{2.5}{68.10}
  & \scalebox{2.5}{82.18}  \\

& \scalebox{4}{\checkmark} & \scalebox{4}{\checkmark}
  & \boldres{\scalebox{2.5}{89.93}}
  & \boldres{\scalebox{2.5}{80.67}}
  & \boldres{\scalebox{2.5}{79.62}}
  & \boldres{\scalebox{2.5}{79.15}}
  & \boldres{\scalebox{2.5}{88.34}}  \\

\bottomrule
\end{tabular}
}
\end{table}

\begin{table}[t]
\caption{Kernel size sensitivity (ACC / F1) anlysis.}
\vspace{-2mm}
\label{tab:kernel}
\centering
\footnotesize
\resizebox{0.46\textwidth}{!}{
\begin{tabular}{lccccccc}
\toprule
\textbf{Dataset} & \textbf{Metric} & \textbf{K=64} & \textbf{K=56} & \textbf{K=48} & \textbf{K=40} & \textbf{K=32}  \\
\midrule
\multirow{2}{*}{\textbf{Neurocon}} 
& \textbf{Accuracy}  & \boldres{87.67}  & 86.61 & 84.56 & 82.22 & 85.61 \\
& \textbf{F1-Score}  & \boldres{82.79}  & 81.90 & 81.11 & 74.39 & 82.24 \\
\midrule
\multirow{2}{*}{\textbf{TaoWu}} 
& \textbf{Accuracy}  & \boldres{86.50}  & 77.50 & 80.00 & 80.00 & 78.00 \\
& \textbf{F1-Score}  & \boldres{86.17}  & 75.77 & 78.95 & 78.92 & 76.25\\
\bottomrule
\end{tabular}
}
\end{table}
\begin{table}[!t]
\centering
\def\arraystretch{1.0}
\caption{Ablation result of the proposed Channel-Independent (CI) strategies \textit{vs.} Channel-Dependent (CD). We design two CD variants: \textit{(i)} CD is introduced by utilizing a channel mixing network to extract inter-ROI interactions before the feature is input into each DeCI block (inter-CD); \textit{(ii)} CD is introduced by utilizing a Transformer Encoder (attention mechanism) to extract inter-ROI interactions on the final output of each ROI (final-CD).}
\vspace{-2mm}
\label{tab:ci_ablation}
\resizebox{0.48\textwidth}{!}{
\begin{tabular}{ccccccc}

    \toprule
    \scalebox{2.5}{\textbf{ }}& \multicolumn{1}{c}{\scalebox{2.5}{\textbf{Design}}} & \multicolumn{1}{c}{\scalebox{2.5}{\textbf{Accuracy
    }}} & \multicolumn{1}{c}{\scalebox{2.5}{\textbf{Precision}}} & \multicolumn{1}{c}{\scalebox{2.5}{\textbf{Recall}}} & \multicolumn{1}{c}{\scalebox{2.5}{\textbf{F1-Score}}} & \multicolumn{1}{c}{\scalebox{2.5}{\textbf{AUROC}}} \\

    \midrule
    \multirow{3}{*}{\scalebox{2.5}{{\textbf{ Mātai}}}}

  &\scalebox{2.5}{inter-CD}
  & \scalebox{2.5}{66.33}
  & \scalebox{2.5}{65.21}
  & \scalebox{2.5}{60.97}
  & \scalebox{2.5}{56.18}
  & \scalebox{2.5}{57.89} \\
  &\scalebox{2.5}{final-CD}
  & \scalebox{2.5}{64.33}
  & \scalebox{2.5}{50.88}
  & \scalebox{2.5}{58.34}
  & \scalebox{2.5}{50.34}
  & \scalebox{2.5}{58.97}  \\
  &\scalebox{2.5}{CI}
  & \boldres{\scalebox{2.5}{73.67}}
  & \boldres{\scalebox{2.5}{73.27}}
  & \boldres{\scalebox{2.5}{70.34}}
  & \boldres{\scalebox{2.5}{67.21}}
  & \boldres{\scalebox{2.5}{66.63}}  \\

\midrule
    \multirow{3}{*}{\scalebox{2.5}{{\textbf{ TaoWu}}}}

  &\scalebox{2.5}{inter-CD}
  & \scalebox{2.5}{73.00}
  & \scalebox{2.5}{78.58}
  & \scalebox{2.5}{73.00}
  & \scalebox{2.5}{70.45}
  & \scalebox{2.5}{68.00}  \\
  &\scalebox{2.5}{final-CD}
  & \scalebox{2.5}{67.50}
  & \scalebox{2.5}{72.90}
  & \scalebox{2.5}{67.50}
  & \scalebox{2.5}{62.55}
  & \scalebox{2.5}{65.75}  \\
  &\scalebox{2.5}{CI}
  & \boldres{\scalebox{2.5}{86.50}}
  & \boldres{\scalebox{2.5}{88.93}}
  & \boldres{\scalebox{2.5}{86.50}}
  & \boldres{\scalebox{2.5}{86.17}}
  & \boldres{\scalebox{2.5}{88.50}}  \\

    \midrule
    \multirow{3}{*}{\scalebox{2.5}{{\textbf{ Neurocon}}}}

  &\scalebox{2.5}{inter-CD}
  & \scalebox{2.5}{76.22}
  & \scalebox{2.5}{67.01}
  & \scalebox{2.5}{68.80}
  & \scalebox{2.5}{64.98}
  & \scalebox{2.5}{68.96}  \\
  &\scalebox{2.5}{final-CD}
  & \scalebox{2.5}{67.22}
  & \scalebox{2.5}{44.06}
  & \scalebox{2.5}{56.47}
  & \scalebox{2.5}{47.73}
  & \scalebox{2.5}{57.91}  \\
  &\scalebox{2.5}{CI}
  & \boldres{\scalebox{2.5}{87.67}}
  & \boldres{\scalebox{2.5}{86.11}}
  & \boldres{\scalebox{2.5}{84.13}}
  & \boldres{\scalebox{2.5}{82.79}}
  & \boldres{\scalebox{2.5}{83.02}}  \\

    \midrule
    \multirow{3}{*}{\scalebox{2.5}{{\textbf{ PPMI}}}}

  &\scalebox{2.5}{inter-CD}
  & \scalebox{2.5}{70.19}
  & \scalebox{2.5}{45.63}
  & \scalebox{2.5}{44.48}
  & \scalebox{2.5}{41.90}
  & \scalebox{2.5}{64.65}  \\
  &\scalebox{2.5}{final-CD}
  & \scalebox{2.5}{63.54}
  & \scalebox{2.5}{28.18}
  & \scalebox{2.5}{29.56}
  & \scalebox{2.5}{25.80}
  & \scalebox{2.5}{54.95}  \\
  &\scalebox{2.5}{CI}
  & \boldres{\scalebox{2.5}{89.93}}
  & \boldres{\scalebox{2.5}{80.67}}
  & \boldres{\scalebox{2.5}{79.62}}
  & \boldres{\scalebox{2.5}{79.15}}
  & \boldres{\scalebox{2.5}{88.34}}  \\

\bottomrule
\end{tabular}
}
\end{table}
\vspace{-10mm}

\begin{figure}[!t]
  \centering
  \includegraphics[width=0.48\textwidth]{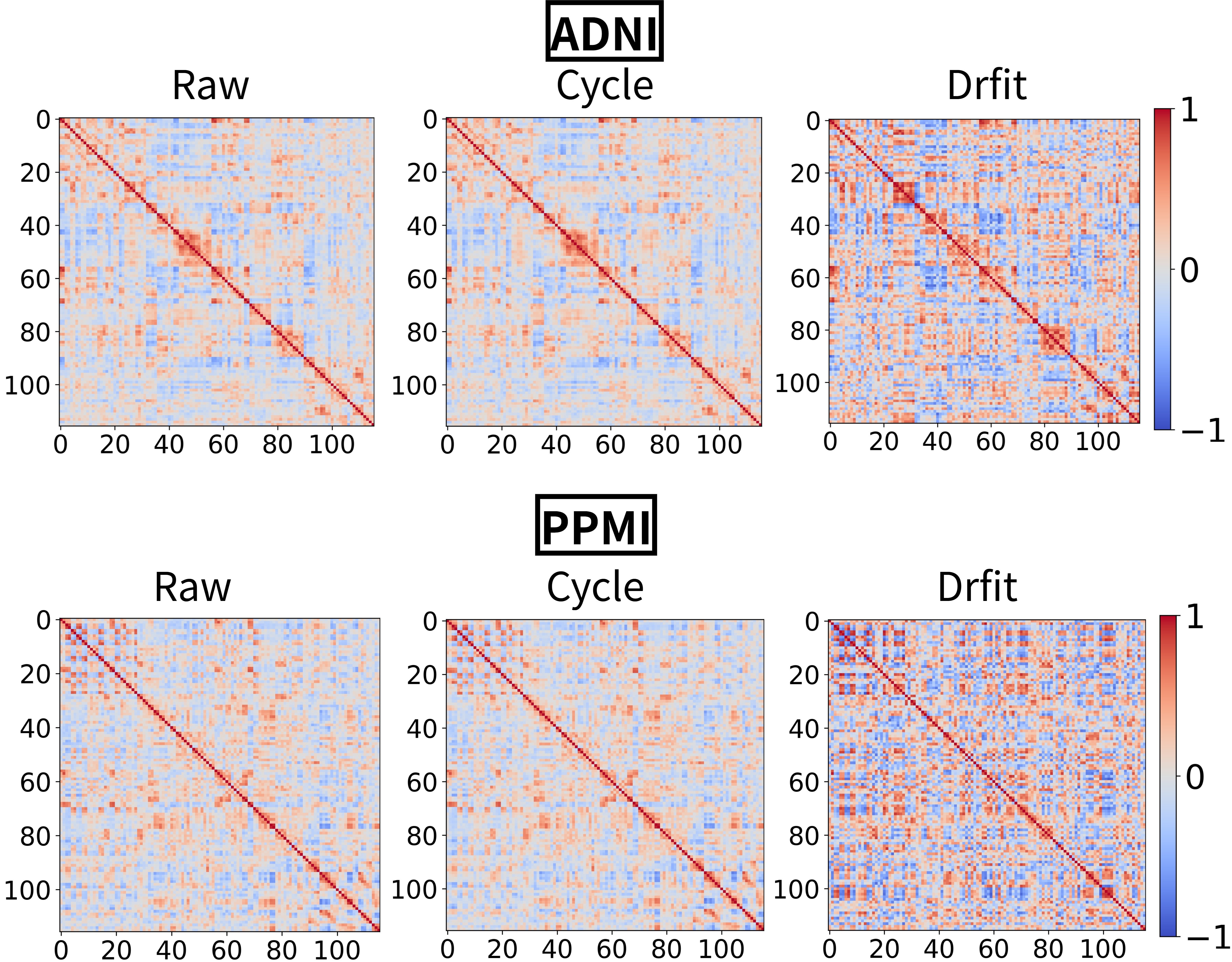}
\caption{Functional connectivity computed using Raw BOLD, Cycle, and Drift components.}
  \label{fig:fc}
\end{figure}

\begin{figure}[!t]
  \centering
  \includegraphics[width=0.48\textwidth]{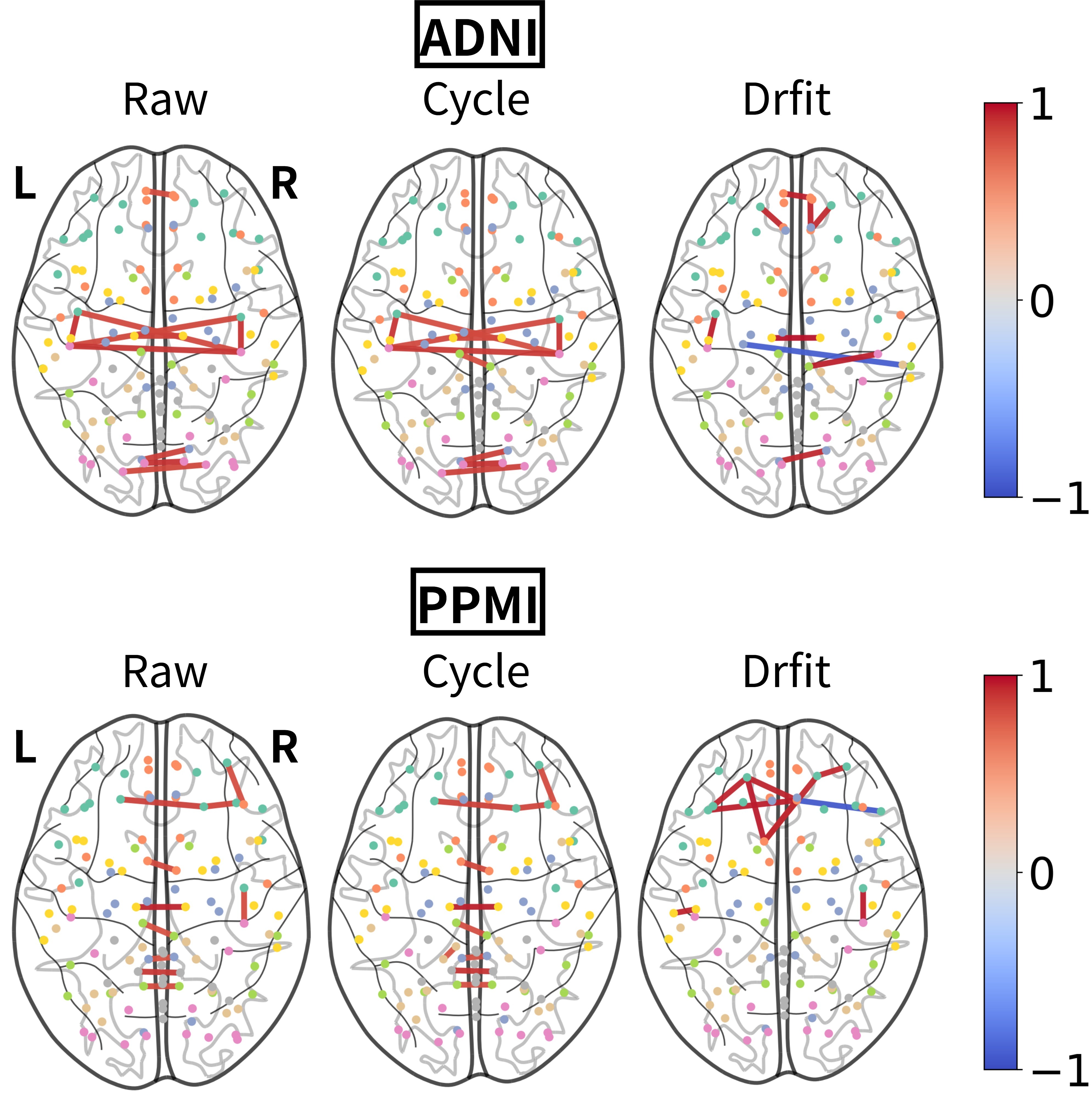}
\caption{Top-10 connectivity from Raw, Cycle, and Drift.}
  \label{fig:roi}
\end{figure}

\begin{table}[!t]
\centering
\caption{Per-ROI variability, reported as the standard deviation (\textit{\textbf{std}}) across all ROIs (\textit{lower is better}).}
\vspace{-2mm}
\label{tab:roi}
\resizebox{0.49\textwidth}{!}{%
\begin{tabular}{lcccccc}
\toprule
\multirow{2}{*}{\scalebox{2.5}{Dataset}} & \multicolumn{3}{c}{\scalebox{2.5}{ADNI}} & \multicolumn{3}{c}{\scalebox{2.5}{PPMI}} \\
\cmidrule(lr){2-4} \cmidrule(lr){5-7}
 & \scalebox{2.5}{Accuracy} & \scalebox{2.5}{F1-Score} & \scalebox{2.5}{AUROC} & \scalebox{2.5}{Accuracy} & \scalebox{2.5}{F1-Score} & \scalebox{2.5}{AUROC}\\
\midrule
\scalebox{2.5}{iTransformer} &  \scalebox{2.5}{3.74}   &  \scalebox{2.5}{2.86}   &  \scalebox{2.5}{2.92}   &  \scalebox{2.5}{5.93}   &  \scalebox{2.5}{6.01}   &  \scalebox{2.5}{4.18}\\
\addlinespace[6pt]
\scalebox{2.5}{DeCI}         &  \scalebox{2.5}{2.76}   &  \scalebox{2.5}{2.75}   &  \scalebox{2.5}{2.37}   &  \scalebox{2.5}{4.65}   &  \scalebox{2.5}{4.94}   &  \scalebox{2.5}{1.81}\\
\bottomrule
\end{tabular}
}
\end{table}

\subsubsection{Neuroscientific Interpretability of DeCI}

Fig.~\ref{fig:fc} provides FC visualization from Raw, Cycle, and Drift. While Cycle closely matches Raw, preserving fast, periodic couplings, Drift reveals slow baseline co-fluctuations and anti-correlations that are weak in Raw. These findings reveal that Cycle and Drift Decomposition reduces temporal mixing in FC and clarifies complementary regimes of coupling. Fig.~\ref{fig:roi} shows Cycle recovering Raw’s focal/short-range links, while Drift represents long-range, slow-varying connections. The non-overlapping edge sets explain why the decomposition adds discriminative information beyond Raw BOLD alone. Table~\ref{tab:roi} reports per-ROI variability of CI vs. CD. DeCI (CI) yields consistently smaller \textbf{std} than the CD baseline (iTransformer), indicating more stable ROI-wise evidence and reduced propagation of spurious inter-ROI dependencies.

\section{Discussion}

\subsubsection{General Time Series Method Triumph Feature Analysis}
\begin{figure}[tp]
  \centering
  \includegraphics[width=0.3\textwidth]{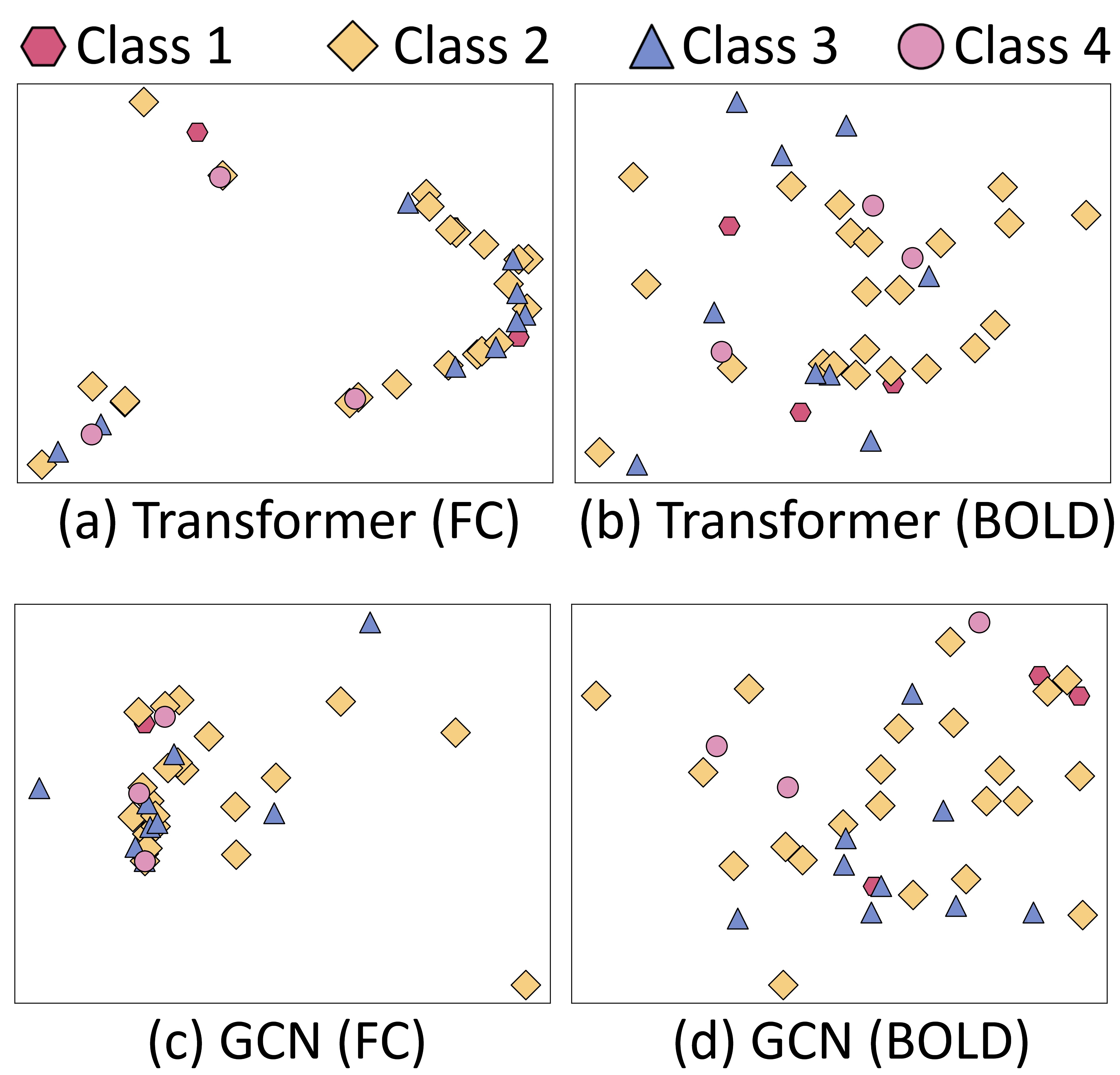}
\caption{Visualization of the learned representations from Transformer and GCN on PPMI, using handcrafted functional connectivity or raw BOLD time series signals, respectively.}
  \label{fig:fc_ts}
\end{figure}

To explain why raw BOLD time series can improve performance, we compare two classic backbone models, Transformer and GCN, using either handcrafted functional connectivity features or raw BOLD signals as input. 
As shown in Fig.\ref{fig:fc_ts}, representations learned from raw BOLD signals display a clearer class separation and more structured organization, while those from FC show considerable class overlap, reducing discriminative power. 
Table~\ref{tab:ts_fc}  provides a quantitative assessment of the class discriminability (clustering) of representations learned by the two models across different inputs. Both models generate representations that achieve superior discriminative (clustering) performance when trained using raw BOLD signals.
\begin{table}[t]
\small
\centering
\caption{Evaluation of clustering performance on PPMI using raw BOLD signals and FC. Unsupervised metric (lower is better): Davies Bouldin Index-DBI. Supervised metrics (after K-Means clustering, higher is better): Normalized Mutual Information-NMI, Homogeneity, and Completeness.}
\vspace{-2mm}
\resizebox{0.38\textwidth}{!}{
\begin{threeparttable}
\begin{tabular}{ccccc}
\midrule
Metrics/Models &\multicolumn{2}{c}{Transformer}&\multicolumn{2}{c}{GCN}\\ 
\cmidrule(l{10pt}r{10pt}){2-3}\cmidrule(l{10pt}r{10pt}){4-5}

 & BOLD & FC & BOLD & FC\\ 
\midrule
DBI & 5.08	&35.52	&3.36	&8.92 \\
NMI	&0.084	&0.062	&0.253	&0.059 \\
Homogeneity	&0.088	&0.061	&0.227	&0.062 \\
Completeness	&0.099	&0.059	&0.215	&0.056 \\
\midrule
\end{tabular}
\end{threeparttable}
}
\label{tab:ts_fc}
\vspace{-2mm}
\end{table}

\subsubsection{Rationale for Channel-Independent strategy}

\textbf{\textit{Our use of \textbf{Channel-Independent (CI)} is a robustness-oriented modeling choice on fMRI signals, rather than a neuroscientific claim that signals from different ROIs are independent.}} Unlike FC/dFC-based methods, which inject inter-ROI interactions at the input level, or Channel-Dependent (CD) temporal models (e.g., iTransformer), which mix ROIs throughout representation learning, DeCI first learns clean \emph{\textbf{per-ROI}} temporal dynamics (e.g., high-frequency Cycle and low-frequency Drift) and only then models their \emph{joint effect} at the decision stage. Concretely, each ROI is treated as a univariate time series to extract its cycle and drift patterns, and the resulting ROI-wise evidence is \emph{jointly fused at the logit level} to produce a subject-level prediction. Without reasonably stable ROI-wise representations, it is difficult to infer reliable inter-ROI interactions; CI therefore prioritizes stabilizing per-ROI dynamics before exploiting inter-regional synergy via logit fusion. In this way, DeCI still captures network-level information through the fusion of ROI-wise logits, rather than through explicit dense feature-level coupling or predefined FC/dFC graphs.

Explicitly modeling pairwise interactions on raw BOLD signals or hidden states is known to amplify motion- and physiology-related artefacts and to overfit spurious correlations in small-sample fMRI~\cite{varoquaux2018fmrici1,birn2006fmrinoise, caballero2017fmrinoise}. From a multivariate time series perspective, CD architectures have higher capacity by directly modeling inter-channel dependencies, but are more prone to overfitting when the channel dimension is high and the samples are limited~\cite{han2023ci1,xu2023maitai,chen2024ci3}. The datasets in this study have $R{=}116$ ROIs and relatively small cohorts, so we adopt CI as a robustness-first inductive bias that reduces reliance on noisy high-dimensional connectivity estimates. Empirically, CI leads to more stable and better-generalizing models, as evidenced by the CI vs. CD ablation (Table~\ref{tab:ci_ablation}) and the robustness study (Fig.~\ref{fig:noise}). Nevertheless, this conservative design may under-represent certain fine-grained network-level interactions; we therefore regard CI as a practical yet restrictive inductive bias and explicitly list this trade-off as a limitation of the current design. Notably, in the only setting where DeCI is not the best (PPMI), the best competitor, PatchTST, is itself a representative CI-based temporal model~\cite{Nie2022patchtst}, which further supports the effectiveness of CI in small-sample fMRI.

\subsubsection{Clinical impact of DeCI}
DeCI can streamline fMRI-based diagnosis along three axes. \textit{(i)} \textbf{\textit{Efficiency}}: DeCI yields linear complexity regarding the number of ROIs, reducing inference time and making same-day analyses more feasible in routine workflows. \textit{(ii)} \textbf{\textit{Robustness}}: By decoupling high-frequency cycles and low-frequency drift under CI, DeCI learns cleaner per-ROI dynamics and shows higher resilience to noise, supporting more reliable diagnosis. \textit{(iii)} \textbf{\textit{Interpretability}}: Cycle/Drift blocks and per-ROI processing naturally expose component-wise and region-wise contributions, enabling clinicians to trace a decision to specific physiological patterns and brain areas—useful for auditability, longitudinal monitoring, and communicating results. Together, these properties position DeCI as a practical, accurate, and explainable engine for future clinical fMRI decision support.

\subsubsection{Clinical Meaning of the Extracted Cycle \& Drift Patterns}
\textit{Unstable Drift aligns with progressive pathology, whereas abnormal (slowed or excessively fragmented) Cycle patterns align with specific cognitive deficits.}
The \textbf{Drift} component captures long-term BOLD variations within the canonical resting-state band that index gradual neurovascular/metabolic shifts accompanying chronic disease burden. In Parkinson’s disease (PD), reduced stability and shorter dwell time in \emph{globally integrated (GI) states}, together with fewer state transitions, are associated with worse motor/cognitive status and dopaminergic loss; thus, a \emph{nonstationary, low-coherence} Drift pattern reflects progression-related changes and can be sensitive to medication state.
The \textbf{\textit{Cycle}} component reflects moment-to-moment coordination among large-scale networks. In PD, alterations in dynamic connectivity between the default-mode and frontoparietal/control networks track attention/executive performance and motor symptoms; accordingly, \emph{higher dFC variability and/or elevated high-frequency fluctuation amplitude} in Cycle patterns can correspond to cognitive deficits and greater symptom severity.

\subsubsection{Technical Limitations} 
\textit{(i) Sensitivity to TR.} The Cycle and Drift Decomposition is frequency-based; different TRs alter the observable low-frequency band and may induce aliasing. We recommend TR-aware (Hz) band definitions and temporal resampling to a common grid.
\textit{(ii) CI-CD trade-off.} CI curbs overfitting and noise propagation, but may underutilize inter-ROI synergies since it only aggregates inter-ROI interaction at the final logit-level. Exploring how to reintroduce inter-ROI interaction (at the input/representation stage) is a promising extension.
\textit{(iii) Overfitting.} Very small datasets pose a higher risk of overfitting (e.g., TaoWu, Neurocon). Though CI can solve this partially, a larger sample size or data augmentation could further alleviate this issue. Additionally, recent advancements have demonstrated that integrating Large Language Models (LLMs) to provide modality-agonistic priors can mitigate overfitting in scarce time series data regimes, which is also a promising avenue.
\textit{(iv) Sequence length.} A short sequence length gives only a limited number of oscillatory cycles, resulting in insufficient frequency resolution. This makes the Cycle and Drift Decomposition in DeCI less stable, and failing in capturing meaningful frequency patterns, which constrains its performance, as shown in Table~\ref{tab:1bench_std} (ABIDE-120), where DeCI delivers a very marginal gain over Leddam on the highly heterogeneous ABIDE dataset.
\textit{(v) Patching strategy.} Though both CI-based, our DeCI noticeably underperforms PatchTST on the PPMI dataset. We hypothesize that this is due to PatchTST tokenizing each ROI into patches, which can better capture fine-grained local patterns~\cite{zhang2024patch,Wangtslb2024}. Such temporal dynamics may be more easily modeled by a patch-based representation. 
\textit{(iv)} and \textit{(v)} indicate that DeCI is not universally optimal across all data characteristics, suggesting that the Decomposition should be chosen cautiously in a data-dependent manner, and combining DeCI with patch-based temporal modeling is a promising direction for future work.

\section{Conclusion}
We benchmark traditional Pearson correlation against state-of-the-art time-series methods, showing that automatic feature learning outperforms handcrafted methods. To improve generalization, we develop DeCI, combining Cycle and Drift Decomposition and Channel-Independence, where Cycle and Drift Decomposition isolates meaningful features and Channel-Independence reduces overfitting. Experimental results on six public fMRI datasets demonstrate DeCI's superior performance, offering a more effective approach for fMRI classification and clinical diagnostics.

\bibliographystyle{ieeetr}
\bibliography{DeCI/DeCI}

\end{document}